\documentclass[sigconf,review=false,anonymous=false, screen]{acmart}
\usepackage{multirow}
\usepackage{graphicx}
\usepackage{caption}
\usepackage{subfigure}
\usepackage{amsthm}
\usepackage{amssymb}

\newtheorem{definition}{Definition}

\newtheorem{example}{Example}
\usepackage{bm}
\usepackage{algorithmic}
\usepackage{algorithm}
\usepackage{color,soul}
\usepackage{adjustbox}

\DeclareMathOperator*{\argmin}{arg\,min}

\AtBeginDocument{%
  \providecommand\BibTeX{{%
    \normalfont B\kern-0.5em{\scshape i\kern-0.25em b}\kern-0.8em\TeX}}}

\copyrightyear{2020}
\acmYear{2020}
\setcopyright{iw3c2w3}
\acmConference[WWW '20]{Proceedings of The Web Conference 2020}{April 20--24, 2020}{Taipei, Taiwan}
\acmBooktitle{Proceedings of The Web Conference 2020 (WWW '20), April 20--24, 2020, Taipei, Taiwan}
\acmPrice{}
\acmDOI{10.1145/3366423.3380087}
\acmISBN{978-1-4503-7023-3/20/04}

\usepackage{hyperref}
\hypersetup{
	colorlinks = true,
    urlcolor  = blue,
	citecolor = {blue}}
\begin{document}

\title{Learning Model-Agnostic Counterfactual Explanations for Tabular Data}





\author{Martin Pawelczyk}
\orcid{1234-5678-9012}
\affiliation{%
  \institution{University of Tuebingen}
  \city{Tuebingen}
  \state{Germany}
  }
\email{martin.pawelczyk@uni-tuebingen.de}

\author{Klaus Broelemann}
\affiliation{%
  \institution{Schufa Holding AG}
  \city{Wiesbaden, Germany}
  }
\email{klaus.broelemann@schufa.de}

\author{Gjergji Kasneci}
\affiliation{%
  \institution{University of Tuebingen}
  \city{Tuebingen}
  \state{Germany}
  }
\email{gjergji.kasneci@uni-tuebinge.de}


\begin{abstract}
Counterfactual explanations can be obtained by identifying the smallest change made to an input vector to influence a prediction in a positive way from a user's viewpoint; for example, from 'loan rejected' to 'awarded' or from 'high risk of cardiovascular disease' to 'low risk'. Previous approaches would not ensure that the produced counterfactuals be \emph{proximate} (i.e., not local outliers) and \emph{connected} to regions with substantial data density (i.e., close to correctly classified observations), two requirements known as \emph{counterfactual faithfulness}. Our contribution is twofold. First, drawing ideas from the manifold learning literature, we develop a framework, called C-CHVAE, that generates \emph{faithful counterfactuals}. Second, we suggest to complement the catalog of counterfactual quality measures using a criterion to quantify the \emph{degree of difficulty} for a certain counterfactual suggestion. Our real world experiments suggest that \emph{faithful counterfactuals} come at the cost of higher \emph{degrees of difficulty}.
\end{abstract}


\maketitle

\section{Introductory remarks}

Machine learning models are increasingly being deployed to automate high-stake decisions in industrial applications, e.g., financial, employment, medical or public services. \citet{wachter2017counterfactual} discuss to establish a legally binding right to request explanations on any prediction that is made based on personal data of an individual. In fact, the EU General Data Protection Regulation (GDPR) includes a right to request ``meaningful information about the logic involved, as well as the significance and the envisaged consequences'' \citep{wachter2017counterfactual} of automated decisions.

As people are increasingly being affected by these automated decisions, it is natural to ask how those affected can be empowered to receive desired results in the future. To this end, \citet{wachter2017counterfactual} suggest using counterfactual explanations. In this context, a counterfactual is defined as a small change made to the input vector to influence a classifier's decision in favor of the person represented by the input vector.

\subsection{A step towards user empowerment}
\paragraph{\textbf{The ``close world'' desideratum}.}
At a high level, \citet{wachter2017counterfactual} formulated the desideratum that counterfactuals should come from a 'possible world' which is 'close' to the user's starting point. \citet{laugel2019issues} formalized the \emph{close world} desideratum and split it into two measurable criteria, \emph{proximity} and \emph{connectedness}. Proximity describes that counterfactuals should not be local outliers and connectedness quantifies whether counterfactuals are close to correctly classified observations. We shortly review both criteria in section \ref{sec:Evaluation_Criteria}. To these two criteria, we add a third one based on percentile shifts of the cumulative distribution function (CDF) of the inputs, as a measure for the \emph{degree of difficulty}. Intuitively, all criteria help quantify how \emph{attainable} suggested counterfactuals are. 
\paragraph{\textbf{The C-CHVAE}}
In this work, our main contribution is a general-purpose framework, the \emph{Counterfactual Conditional Heterogeneous Autoencoder}, C-CHVAE, which allows finding (multiple) counterfactual feature sets while generating counterfactuals with high occurrence probability. This is a fundamental requirement towards \emph{attainability} of counterfactuals. In particular, our framework is compatible with a multitude of autoencoder (AE) architectures as long as the AE allows both modelling of heterogeneous data and approximating the conditional log likelihood of the the mutable/free inputs given the immutable/protected ones. Moreover, the C-CHVAE does not require access to a distance function (for the input space) and is classifier agnostic. Part of this work was previously published as a \emph{NeurIPS HCML workshop paper} \cite{pawelczyk2019user}. Our source code can be found at: \url{https://github.com/MartinPawel/c-chvae}.

\subsection{Challenges for counterfactuals}
\paragraph{\textbf{Attainability}.} Intuitively, a counterfactual is attainable, if it is jointly (1) a 'close' suggestion that is not a local outlier, (2) similar to correctly classified observations and (3) associated with low total CDF percentile shifts. Hence, in our point of view, \emph{attainability} is a composition of faithful counterfactuals ((1) and (2)) which are at the same time not too difficult to attain (3). To reach a better understanding, let us translate conditions (1), (2) and (3) into the following synthetic bank loan setting: a client applies for a loan and a bank employs a counterfactual empowerment tool. Under these circumstances, we focus on one problematic aspect. The tool could make suggestions that '\emph{lie outside of a client's wheelhouse}', that is to say, it is not reasonable to suggest counterfactuals that (a) one would typically not observe in the data, (b) that are not typical for the subgroup of users the client belongs to, and that (c) are extremely difficult to attain, where difficulty is measured in terms of the percentiles of the CDF of the given inputs. For example, in table \ref{tab:HpyotheticalCounterfactual}, the suggestion made by the second method is likely not attainable given her age and education level.
\begin{table*}
    \centering
    \begin{tabular}{cclcccccccc}
        \toprule
         Method & ID & Input subset & Current & Percentile & Counterfactual & Percentile & Shift & Tot. shift & L. Outlier & Connected \\
         \cmidrule(lr){1-1}  \cmidrule(lr){2-2}  \cmidrule(lr){3-3} \cmidrule(lr){4-5} \cmidrule(lr){6-7} \cmidrule(lr){8-11}
         \multirow{2}{*}{I} & \multirow{2}{*}{1} & \small{\texttt{credit card debt}} & 5000 & $\bm{55}$ &  3500 & $\bm{75}$ & $\bm{20}$ & \multirow{2}{*}{40} & \multirow{2}{*}{No} & \multirow{2}{*}{Yes}  \\
         & & \small{\texttt{saving account}} & 200 & $\bm{45}$ & 600 & $\bm{65}$ & $\bm{20}$ & & & \\
         \cmidrule(lr){1-11}
         \multirow{2}{*}{II} & \multirow{2}{*}{1} & \small{\texttt{monthly income (\$)}} & 2500 & $\bm{40}$ &  10000 & $\bm{95}$ & $\bm{55}$ & \multirow{2}{*}{75} & \multirow{2}{*}{Yes} & \multirow{2}{*}{No} \\
         & &\small{\texttt{\# loans elsewhere}} & 5 & $\bm{85}$ & 2 & $\bm{65}$ & $\bm{20}$ & & & \\
         \bottomrule
    \end{tabular}
    \caption{\textbf{Hypothetical counterfactuals} for the same 22 year old individual without a college degree, who was denied credit. Suggestions were made by two different methods for a given classifier $f$. The rows suggest how a subset of free inputs would need to change to obtain credit, i.\ e. from $\hat{y}=0$ to $\hat{y}=1$. For the first empowerment technology, the suggestion might be reasonable whereas for the second one, the suggestion looks atypical and could be difficult to attain measured in terms of connectedness to existing knowledge, an outlier measure and the total percentile shift.}
    \label{tab:HpyotheticalCounterfactual}
\end{table*}

\paragraph{\textbf{Similarity via latent distance}.}
Additionally, in health, banking or credit scoring contexts we often face continuous, ordinal and nominal inputs concurrently. This is also known as \emph{heterogeneous} or \emph{tabular} data. For this type of data, it can sometimes be difficult to measure distance in a meaningful way (e.\ g. measuring distance between different occupations). Furthermore, existing methods leave the elicitation of appropriate distance/cost functions up to (expert) opinions \citep{wachter2017counterfactual,grath2018interpretable,laugel2017inverse,lash2017generalized,spangher2018actionable}, which can vary considerably across individuals \citep{grgic2018human}. Therefore, we suggest measuring similarity between the input feature $\bm{x}_i$ and a potential counterfactual $\tilde{\bm{x}}_i$ as follows.

\begin{definition}[Latent distance]
Let $\bm{x}_i, \bm{x}_j \in \mathbb{R}^n$ be two observations in input space with corresponding lower dimensional representations $\bm{z}_i,\bm{z}_j\in \mathbb{R}^k \text{ with } k<n$, in latent space. Then the distance $d_L(\bm{x}_i,\bm{x}_j):=\lVert \bm{z}_i-\bm{z}_j \rVert_p$ is called the latent distance of $\bm{x}_i$ and $\bm{x}_j$.
\label{def:latent-distance}
\end{definition}


\subsection{Overview}
\begin{figure}
    \centering
    \includegraphics[scale=0.56]{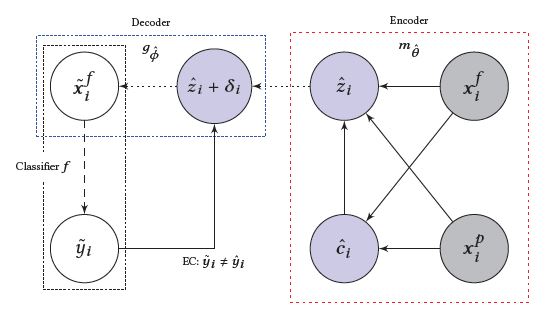}

    \caption{\textbf{Autoencoding Counterfactual search}. The learned encoder, $m_{\hat{\theta}}$, maps heterogeneous protected and free features, $\bm{x}^p$ and $\bm{x}^f$, and latent mixture components, $\hat{\bm{c}}$, into a latent representation, $\hat{\bm{z}}$. The learned decoder, $g_{\hat{\phi}}$, reconstructs the free inputs $\bm{x}^f$ from the perturbed representation, providing a potential counterfactual, $\tilde{\bm{x}}=(\bm{x}^p,\tilde{\bm{x}}^f)$. The counterfactual acts like a typical observation from the data distribution. Next, we feed the potential counterfactual $\tilde{\bm{x}}$ to the classifier, $f$. We stop the search, if the EC condition is met.}
    \label{fig:CounterfactualSearch}
\end{figure}

\paragraph{\textbf{Learning faithful counterfactuals via the C-CHVAE}}
We suggest embedding counterfactual search into a data density approximator, here a variational autoencoder (VAE) \citep{kingma2013auto}. The idea is to use the VAE as a \emph{search device} to find counterfactuals that are \emph{proximate} and \emph{connected} to the input data. The intuition of this approach becomes apparent by considering each part of the VAE in turn. As opposed to classical generative model contexts, the encoder part is not discarded at \emph{test time/generation time}. Indeed, it is the \emph{trained encoder} that plays a crucial role: given the original heterogeneous data, the encoder specifies a lower dimensional, real-valued and dense representation of that data, $\bm{z}$, Therefore, it is the encoder that determines which low-dimensional neighbourhood we should look to for potential counterfactuals. Next, we perturb the low dimensional data representation, $\bm{z} + \bm{\delta}$, and feed the perturbed representation into the decoder. For small perturbations the decoder gives a potential counterfactual by reconstructing the input data from the perturbed representation. This counterfactual is likely to occur. Next, the potential counterfactual is passed to the pretrained classifier, which we ask whether the prediction was altered. Figure \ref{fig:CounterfactualSearch} represents this mechanism.



\paragraph{\textbf{Consistent search for heterogeneous data}.}
While we aim to avoid altering immutable inputs, such as \emph{age} or \emph{education}, it is reasonable to believe that the immutable inputs can have an impact on what is attainable to the individual. Thus, the immutable inputs should influence the neighbourhood search for counterfactuals. For example, certain drugs can have different treatment effects, depending on whether a patient is male or female \cite{regitz2012sex}. Hence, we wish to generate \emph{conditionally consistent} counterfactuals. 

Again, consider Figure \ref{fig:CounterfactualSearch} for an intuition of counterfactual search in the presence of immutable inputs. Unlike in vanilla VAEs, we assume a Gaussian mixture prior on the latent variables where each mixture component is also estimated by the immutable inputs. This helps cluster the latent space and has the advantage that we look for counterfactuals among \emph{semantically similar} alternatives.

\paragraph{\textbf{Contribution}.}
The C-CHVAE is a general-purpose framework that generates counterfactuals. Its main merits are:
\begin{itemize}
    \item \textbf{Faithful counterfactuals.} The generated counterfactuals are \emph{proximate} and \emph{connected} to regions of high data density and therefore likely attainable, addressing the most important desiderata in the literature on counterfactuals \cite{wachter2017counterfactual,laugel2019issues}; 
    \item \textbf{Suitable for tabular data and classifier agnostic.} The data distribution is modelled by an autoencoder that handles heterogeneous data and interval constraints by choosing appropriate likelihood models. It can also be combined with a multitude of autoencoder architectures \citep{kingma2013auto,sohn2015learning,nazabal2018handling, makhzani2015adversarial,tolstikhin2017wasserstein,ivanov2018variational};
    \item \textbf{No ad-hoc distance measures for \emph{input} data.} The C-CHVAE does not require ad-hoc predefined distance measures for input data to generate counterfactuals. This is can be an advantage over existing work, since it can be difficult to devise meaningful distance measures for tabular data.
\end{itemize}



\section{Related Literature}\label{sec:related_lit}
 \paragraph{\textbf{Explainability through counterfactuals}}
 At a meta level, the major difference separating our work from previous approaches is that we \emph{learn} a separate model to learn similarity in latent space and use this model to generate counterfactual recommendations. Doing this allows us to generate counterfactuals that lie on the data manifold.
 
 Approaches dealing with heterogeneous data rely on integer programming optimization \citep{spangher2018actionable,russell2019efficient}. To produce counterfactuals that take on reasonable values (e.\ g. non negative values for wage income) one directly specifies the set of features and their respective support subject to change. The C-CHVAE also allows for such constraints by choosing the likelihood functions for each feature appropriately (see Section \ref{sec:CHVAE} and our github repo.). 
 
A closely related collection of approaches assumes that distances or costs between any two points can be measured in a meaningful way \citep{wachter2017counterfactual,lash2017generalized,laugel2017inverse,grath2018interpretable,spangher2018actionable}. The C-CHVAE, however, does not rely on task-specific, predefined similarity functions between the inputs and counterfactuals. For a \emph{given autoencoder architecture}, we learn similarity between inputs and counterfactuals from the data.

Other approaches strongly rely on the pretrained classifier $f$ and make use of restrictive assumptions, e.\ g. that $f$ stems from a certain hypothesis class. For example, \citet{spangher2018actionable} and \citet{tolomei2017interpretable} assume the pretrained classifiers to be linear or tree based respectively, which can restrict usefulness.

In independent work from our's, \citet{joshi2019towards} suggest a similar explanation model, however, they focus on causal models and are less concerned with the issue of evaluating counterfactual explanations.



\paragraph{\textbf{Adversarial perturbations}} 
Since counterfactuals are often generated independently of the underlying classification model, they are related to universal adversarial attacks (see for example \citet{brown2017adversarial}). While adversarial examples aim to alter the prediction a deep neural network makes on a data point via small and \emph{imperceptible} changes, counterfactuals aim to alter data points to suggest \emph{impactful} changes to individuals. Notice that counterfactuals do not fool a classifier in a classical sense, since individuals need to \emph{exert real-world effort} to achieve the desired prediction. Since a review of the entire literature on adversarial attacks goes beyond the scope of this work, we refer the reader to the survey by \citet{akhtar2018threat}. 
For an overview of counterfactual generation methods consider table \ref{tab:lit_overview}.


\begin{table}
    \centering
    \begin{tabular}{ccccc}
        \toprule
        \multirow{2}{*}{Method} & \multirow{2}{*}{Train} & \multirow{2}{*}{ \shortstack[c]{ Classifier \\ agnostic}} & \multirow{2}{*}{Classifier} & \multirow{2}{*}{ \shortstack[c]{ Tabular \\ data} } \\
         \\
         \cmidrule(lr){1-1} \cmidrule(lr){2-5}
         AR \citep{spangher2018actionable} & No & No & Lin. Models & No \\
         HCLS \citep{lash2017generalized} & No & No & SVM & No  \\
         GS \citep{laugel2017inverse} & No & Yes & All & No  \\
         FT \citep{tolomei2017interpretable} & No & No & Trees & No  \\
         \textbf{C-CHVAE} (ours) & Yes & Yes & All & Yes  \\
         \bottomrule
    \end{tabular}
    \caption{Overview of existing counterfactual generation methods. `Train' indicates that a method requires training. `Classifier agnostic' means whether a method can be combined with any black-box classifier.}
    \label{tab:lit_overview}
\end{table}


\paragraph{\textbf{Notation}.} In the remainder of this work, we denote the $D$ dimensional feature space as $\mathcal{X}=\mathbb{R}^D$ and the feature vector for observation $i$ by $\bm{x}_i \in \mathcal{X}$. We split the feature space into two disjoint feature subspaces of immutable (i.\ e. protected) and free features denoted by $\mathcal{X}_p = \mathbb{R}^{D_p}$ and $\mathcal{X}_f = \mathbb{R}^{D_f}$ respectively such that w.l.o.g. $\mathcal{X} = \mathcal{X}_p \times \mathcal{X}_f$ and $\bm{x}_i=(\bm{x}_i^p, \bm{x}_i^f)$. This means in particular that the $d$-th free feature of $x_i$ is given by $x^f_{d,i}=x_{d,i}$ and the $d$-th protected feature is given by $x^p_{d,i}=x_{d+D_f,i}$. Let $\bm{z}\in\mathcal{Z}=\mathbb{R}^k$ denote the latent space representation of $\bm{x}$. The labels corresponding to the $i$'th observation are denoted by $y_i \in \mathcal{Y} = \{0,1\}$. Moreover, we assume a given pretrained classifier $f: \mathcal{X} \xrightarrow{} \mathcal{Y}$. Further, we introduce the following sets: $H^- = \{\bm{x} \in \mathcal{X}: f(\bm{x}) = 0 \}, H^+ = \{\bm{x} \in \mathcal{X}: f(\bm{x}) = 1  \}, D^+ = \{\bm{x}_i \in \mathcal{X}: y_i = 1 \}$. We attempt to find an explainer $E:\mathcal{X} \xrightarrow{ }\mathcal{X}$, generating counterfactuals $E(\bm{x}) = \tilde{\bm{x}}$, such that $f(\bm{x}) \neq f(E(\bm{x}))$. Finally, values with $\hat{\cdot}$ usually denote estimated quantities, values carrying $\tilde{\cdot}$ denote candidate values and values with $\cdot^*$ denote the best value among a number of candidate values.

\section{Background}\label{sec:Background}

\subsection{(Conditional) Variational Autoencoder}
The simple VAE is often accompanied by an isotropic Gaussian prior $p(\bm{z}) = \mathcal{N}(\bm{0},\bm{I})$. We then aim to optimize the following objective known as the Evidence Lower Bound (ELBO),
\begin{equation*}
\begin{split}
    \mathrm{L}_{VAE}(p,q) = &\mathbb{E}_{q(\bm{z}|\bm{x}^f)} [\log p(\bm{x}^f|\bm{z})] -  \mathrm{D}_{KL}[q(\bm{z}|\bm{x}^f)\vert \vert p(\bm{z})].
\end{split}
\end{equation*}
This objective bounds the data log likelihood, $\log~ p(\bm{x}^f)$, from below. In the simple model, the decoder and the encoder are chosen to be Gaussians, that is, $q(\bm{z}|\bm{x}^f) = \mathcal{N}(\bm{z}|\bm{\mu}_q, \bm{\Sigma}_q)$ and $p(\bm{x}^f|\bm{z}) = \mathcal{N}(\bm{x}^f|\bm{\mu}_p, \bm{\Sigma}_p)$, where the distributional parameters $\bm{\mu}(\cdot)$ and $\bm{\Sigma}(\cdot)$ are estimated by neural networks. If all inputs were binary instead, one could use a Bernoulli decoder, $p(\bm{x}^f|\bm{z})=Ber(\bm{x}^f|\varrho_p(\bm{z}))$.

Conditioning on a set of inputs, say $\bm{x}^p$, the objective that bounds the conditional log likelihood, $\log~ p(\bm{x}^f|\bm{x}^p)$, can be written as \citep{sohn2015learning},
\begin{equation}
\begin{split}
            \mathrm{L}_{CVAE}(p,q) = & \mathbb{E}_{q(\bm{z}|\bm{x}^f, \bm{x}^p)} [\log p(\bm{x}^f|\bm{z}, \bm{x}^p)] - \\ & \mathrm{D}_{KL}[q(\bm{z}|\bm{x}^f,\bm{x}^p)\vert \vert p(\bm{z}|\bm{x}^p)],
    \label{eq:vanilla_cvae}
\end{split}
\end{equation}
where one assumes that the prior $p(\bm{z}|\bm{x}^p)$ is still an isotropic Gaussian, i.e. $\bm{z}|\bm{x}^p \sim \mathcal{N}(\bm{0}, \bm{I})$. We will refer to this model as the CVAE.

\section{C-CHVAE} \label{sec:CCHVAE}
In this part, we present both our objective function and the CHVAE architecture in Sections \ref{sec:counterfactual_search_objective} and \ref{sec:CHVAE}, respectively.

\subsection{The C- in C-CHVAE}\label{sec:counterfactual_search_objective}
We take the pretrained, potentially non-linear, classifier $f(\cdot)$ as given, which can also be a training time fairness constraint classifier \citep{zafar2017fairness,agarwal2018reductions}. Let us denote the encoder function, parameterized by $\theta$, by $m_{\theta}(\cdot;\bm{x}^p))$, taking arguments $\bm{x}^f$. The decoder function, parametrized by $\phi$, is denoted by $g_{\phi}(\cdot)$. It has inputs $m_{\theta}(\bm{x}^f; \bm{x}^p) = ~ \bm{z} \in  \mathbb{R}^k$. Then our objective reads as follows,
\begin{align}
    &\underset{\delta \in \mathcal{Z}}{\text{min}}  ~ ||\bm{\delta}||~ \text{subject to} \label{eq:MyProblem} \\
    \begin{split}
        & f(g_{\hat{\phi}}(m_{\hat{\theta}}(\bm{x})+\bm{\delta}), \bm{x}^p) \neq f(\bm{x}^f,\bm{x}^p) \label{eq:Classification} ~ \&     \end{split}
 \\
    & \underset{\phi, \theta}{\text{min}} ~ \ell(\bm{x}, g_{\phi}(m_{\theta}(\bm{x}^f;\bm{x}^p))) - \Omega(m_{\theta}(\bm{x}^f;\bm{x}^p)), \label{eq:Embedding}
\end{align}
where $\Omega(\cdot)$ is a regularizer on the latent space and $\lVert \cdot \rVert$ denotes the p-norm.
The idea behind the objective is as follows. First, \eqref{eq:Embedding} approximates the conditional log likelihood, $p(\bm{x}^f|\bm{x}^p)$, while learning a lower dimensional latent representation. Subsequently, we use this latent representation, $\hat{\bm{z}}=  m_{\hat{\theta}}(\bm{x}^f; \bm{x}^p)$, to search for counterfactuals (\eqref{eq:MyProblem} and \eqref{eq:Classification}). If the perturbation on $\hat{\bm{z}}$ is small enough, the trained decoder $g_{\hat{\phi}}$ gives a reconstruction $\tilde{\bm{x}}^{f*}$ that (a) is similar to $\bm{x}^{f}$, (b) satisfies the empowerment condition \eqref{eq:Classification}, and (c) lies in regions where we would usually expect data. Also, notice that this regularizer effectively plays the role of the distance function. It determines the neighbourhood of $\bm{x}$ in which we search for counterfactuals.

\subsection{CHVAE} \label{sec:CHVAE}
To solve the above optimization problem defined in \eqref{eq:MyProblem}-\eqref{eq:Embedding}, it is crucial to elicit an appropriate autoencoder architecture. We adjust the HVAE \citep{nazabal2018handling} so that it approximates conditional densities.

\paragraph{\textbf{Factorized decoder}.}
We suggest using the following hierarchical model to accommodate the generation of counterfactuals conditional on some immutable attributes. The factorized decoder with a conditional uniform Gaussian mixture prior (\eqref{eq:cond_mix_prior_one} and \eqref{eq:cond_mix_prior_two}) with parameters $\pi_l=1/L$ for all mixture components $l$ reads:
\begin{align}
p(\bm{c}_i|\bm{x}_{i}^p) &= Cat(\bm{\pi}) \label{eq:cond_mix_prior_one} \\
p(\bm{z}_i|\bm{c}_{i}, \bm{x}_{i}^p) &= \mathcal{N}(\bm{\mu}_p(\bm{c}_i),\bm{I}_K) \label{eq:cond_mix_prior_two}  \\
p(\bm{z}_i, \bm{x}^f_{i}, \bm{c}_i|\bm{x}_{i}^p) &= p(\bm{z}_i, \bm{c}_i|\bm{x}_{i}^p) ~ \prod_{d = 1}^{D_f} ~ p(\bm{x}^f_{d,i}|\bm{z}_i,\bm{c}_i,\bm{x}_{i}^p) \nonumber  \\  
\begin{split} &= p(\bm{z}_i| \bm{c}_i, \bm{x}_{i}^p) p(\bm{c}_i|\bm{x}_{i}^p)  \cdot \prod_{d = 1}^{D_f} p(\bm{x}^f_{d,i}|\bm{z}_i,\bm{c}_i, \bm{x}_{i}^p), 
\end{split}
\label{eq:Factorization}
\end{align}
where $\bm{z}_i \in \mathbb{R}^k$ is the continuous latent vector and $\bm{c}_{i} \in \mathbb{R}^C$ is a vector indicating mixture components, generating the instance $\bm{x}^f_{i} \in \mathbb{R}^{D_f}$. Note that 
\eqref{eq:cond_mix_prior_one} and \eqref{eq:cond_mix_prior_two}, where we assume independence between $\bm{c}_i$ and $\bm{x}_i^p$, are analogous to the prior on $\bm{z}$ in the CVAE above, \eqref{eq:vanilla_cvae}. Moreover, the intuition behind the mixture prior is to facilitate clustering of the latent space in a meaningful way.

Since the factorized decoder in \eqref{eq:Factorization} is a composition of various likelihood models, we can use one likelihood function per input, giving rise to modelling data with real-valued, positive real valued, count, categorical and ordinal values, concurrently. Additionally, the modelling framework lets us specify a variety of interval constraints by choosing likelihoods appropriately (e.g.\ truncated normal distribution or Beta distribution for interval data). 

\paragraph{\textbf{Factorized encoder}.}
Then the factorized encoder is given by:
\begin{align}
q(\bm{c}_i|\bm{x}^p_{i},\bm{x}^f_{i}) &= Cat(\bm{\pi}(\bm{x}^p_{i},\bm{x}^f_{i})) \nonumber  \\
q(\bm{z}_i|\bm{x}^f_{i}, \bm{x}^p_{i}, \bm{c}_{i}) &= \mathcal{N}(\bm{\mu}_q(\bm{x}^f_{i}, \bm{x}^p_{i}, \bm{c}_i),\bm{\Sigma}_q(\bm{x}^f_{i},\bm{x}^p_{i},\bm{c}_i)) \nonumber  \\
q(\bm{z}_i, \bm{x}^f_{i}, \bm{c}_i|\bm{x}^p_{i}) &= q(\bm{z}_i, \bm{c}_i|\bm{x}^p_{i}) ~ \prod_{d = 1}^{D_f} ~ p(\bm{x}^f_{d,i}|\bm{z}_i,\bm{c}_i, \bm{x}^p_{i}) \nonumber  \\  
\begin{split} &= q(\bm{z}_i| \bm{c}_i,\bm{x}^p_{i}) q(\bm{c}_i|\bm{x}^p_{i}) \cdot \prod_{d = 1}^{D_f} p(\bm{x}^f_{d,i}|\bm{z}_i,\bm{c}_i, \bm{x}^p_{i}). \label{eq:factorization_enc} \end{split}
\end{align}

\paragraph{\textbf{Parameter sharing and likelihood models}.}
Unlike in the vanilla CVAE in \eqref{eq:vanilla_cvae}, which is only suitable for one data type at the time, the decoder was factorized into multiple likelihood models. In practice, one needs to carefully specify one likelihood model per input dimension $p(\bm{x}^f_{d,i}|\bm{z}_i, \bm{c}_i)$. In our github repository, we describe more details of the model architecture and which likelihood models we have chosen.


\paragraph{\textbf{ELBO}}
The evidence lower bound (ELBO) can be derived as:
\begin{align*}
\begin{split} & \log p(\bm{x}^f|\bm{x}^p)  \geq \mathbb{E}_{q(\bm{c}_i,\bm{z}_i|\bm{x}^f_i, \bm{x}^p_i)}\sum_{d=1}^{D_f} \log p(\bm{x}^f_{d,i}|\bm{z}_i, \bm{c}_i, \bm{x}_i^p) \\ & -\sum_i \mathbb{E}_{q(\bm{c}_i|\bm{x}_i^f,\bm{x}_i^p)} \mathrm{D}_{KL}[q(\bm{z}_i|\bm{c}_i,\bm{x}^f_i,\bm{x}^p_i)||p(\bm{z}_i|\bm{c}_i, \bm{x}^p_i)] \\ & - \sum_i  \mathrm{D}_{KL}[q(\bm{c}_i|\bm{x}_i^f,\bm{x}_i^p)||p(\bm{c}_i|\bm{x}_i^p)]
\end{split}
\end{align*}
where we recognize the influence of the factorized decoder in the first line, effectively allowing us to model complex, heterogeneous data distributions.

\subsection{Counterfactual search algorithm} \label{sec:algorithm}
As inputs, our algorithm requires any pretrained classifier $f$ and the trained decoder and encoder from the CHVAE. It returns the closest $E(\bm{x})$ due to a nearest neighbour style search in the latent space. The details can be found in our github repository, but it uses a standard procedure to generate random numbers distributed uniformly over a sphere \citep{harman2010decompositional,laugel2017inverse} around the latent observation $\hat{\bm{z}}$. Thus, we sample observations $\tilde{\bm{z}}$ in $l_p$-spheres around the point $\hat{\bm{z}}$ until we find a counterfactual explanation $\tilde{\bm{x}}^*$. 

\section{Evaluating attainability of counterfactuals}\label{sec:Evaluation_Criteria}
To quantify faithfulness, \citep{laugel2019issues} suggest two measures, which we shortly review here since they do not belong to the catalog of commonly used evaluation measures (such as for example accuracy). Their two suggested measures quantify \emph{proximity} (i.e. whether $E(\bm{x})$ is a local outlier) and \emph{connectedness} (i.e. whether $E(\bm{x})$ is connected to other correctly classified observations from the same class). However, these measures do not indicate the \emph{degree of difficulty} for the individual to attain a certain counterfactual given the current state. We suggest two appropriate measures in \ref{sec:difficulty}.

\subsection{Counterfactual faithfulness}\label{sec:faithfulness}
\paragraph{\textbf{Proximity}.} 

Ideally, the distance between a counterfactual explanation $E(\bm{x})$ and its closest, non-counterfactual neighbour $\bm{a}_0 \in H^+ \cap D^+$ should be small:
  \begin{equation*}
    \bm{a}_0 = \underset{\bm{x} \in H^+ \cap D^+}{\argmin}~  d(E(\bm{x}),\bm{x}).
\end{equation*}
Moreover, it is required that the observation resembling our counterfactual, $\bm{a}_0$, be close to the rest of the data, which gives rise to the following relative metric:
\begin{equation*}
    P(E(\bm{x})) = \frac{d(E(\bm{x}), \bm{a}_0)}{\underset{\bm{x} \neq \bm{a}_0 \in H^+ \cap D^+}{\min}~d(\bm{a}_0 ,\bm{x})}.
    \label{eq:LOF}
\end{equation*}
The intuition behind this measure is to help evaluate whether counterfactuals are outliers relative to correctly classified observations. 



\paragraph{\textbf{Connectedness}.} We say that that a counterfactual $\bm{e}$ and an observation $\bm{a}$ are $\epsilon$-chained, with $\epsilon >0$, if there exists a sequence $\bm{e}_0, \bm{e}_1, ..., \bm{e}_N \in \mathcal{X}$ such that $\bm{e}_0 = \bm{e}, \bm{e}_N = \bm{a}$ and $\forall i < N$, $d(\bm{e}_i, \bm{e}_{i+1}) < \epsilon$ and $f(\bm{e}) = f(\bm{a})$. Now, given an appropriate value for $\epsilon$, we can evaluate the connectedness of a counterfactual $E(\bm{x})$ using a binary score: $C(E(\bm{x})) = 1$, if $E(\bm{x})$ is $\epsilon$-connected to $\bm{a} \in H^+ \cap D^+$ and $C(E(\bm{x})) = 0$, otherwise.



\subsection{Degree of difficulty} \label{sec:difficulty}
\paragraph{\textbf{Individual costs of counterfactuals}.}
We suggest to measure the degree of difficulty of a certain counterfactual suggestion $\tilde{\bm{x}}$ in terms of the percentiles of $x^f_d = \{x_{i,d}\}_{i=1}^N$ and $\tilde{x}_d^{f*}$: $Q_j(\tilde{x}_d^{f*})$ and $Q_d(x^f_d)$ where $Q_d(\cdot)$ is the cumulative density function of $x^f_d$. As an example, a cost of $p$ suggests changing a free feature by at least $p$ percentiles to receive a desired result.

 We suggest two measures with the following properties: (a) $cost(x^f_d;x^f_d) = \bm{0}_{N \times 1}$, implying that staying at the current state is costless and (b) $cost(\tilde{x}^f + \nu \bm{1}_{N \times 1} ;x^f) \geq cost(\tilde{x}^f ;x^f) \text{ with } \nu \geq 0$, that is, the further from the current state, the more difficulties we have to incur to achieve the suggestion. The \emph{difficulty measures} then read as follows:
\begin{align}
    cost_{1}(\tilde{x}^{f*};x^f) &= \sum_{d=1}^{D_f} |(Q_d(\tilde{x}_d^{f*})-{Q_d(x_d^f)}|, \label{eq:total_shift} \\
    cost_{2}(\tilde{x}^{f*};x^f) &= \underset{d}{\max}~ |Q_d(\tilde{x}_d^{f*})-Q_d(x_d^f)|. \label{eq:max_shift}
\end{align}  
The total percentile shift (TS) in \eqref{eq:total_shift} can be thought of as a baseline measure for how attainable a certain counterfactual suggestion might be. The maximum percentile shift (MS) in \eqref{eq:max_shift} across all free features reflects the maximum difficulty across all mutable features. 

\section{Experiments}\label{sec:experiments}
\subsection{Synthetic experiments}
\begin{figure}
\centering
\begin{subfigure}[Reconstructed train data generated by different {$\hat{z}$} (coloured). \label{fig:Recon_Blobs}]{
\includegraphics[width=0.30\columnwidth]{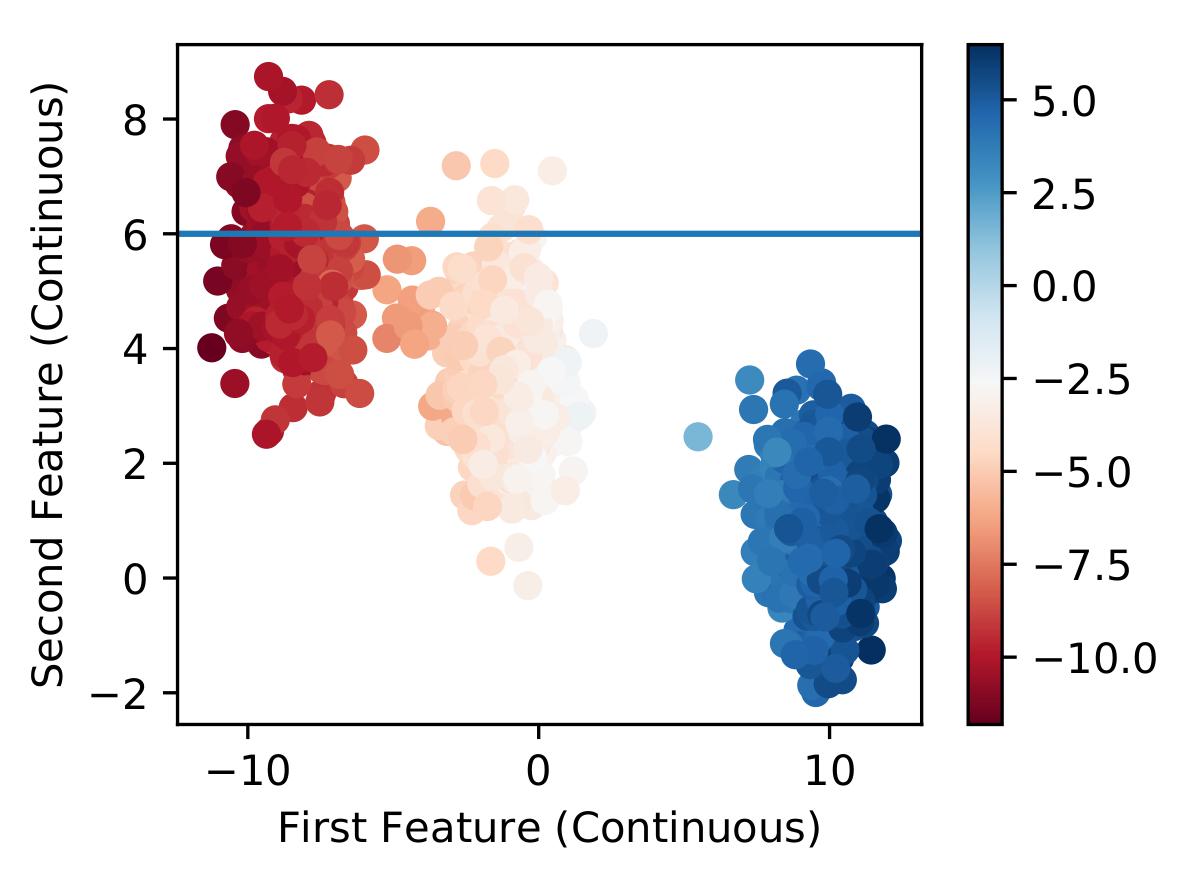}}
\end{subfigure}\hfill
\begin{subfigure}[Density of {$\hat{\bm{z}}$}. Colours aligned so that left red {$\hat{z}$} generates left {$\hat{x}$} in \ref{fig:Recon_Blobs}. \label{fig:ThreeMode_Blobs}]{
\includegraphics[width=0.30\columnwidth]{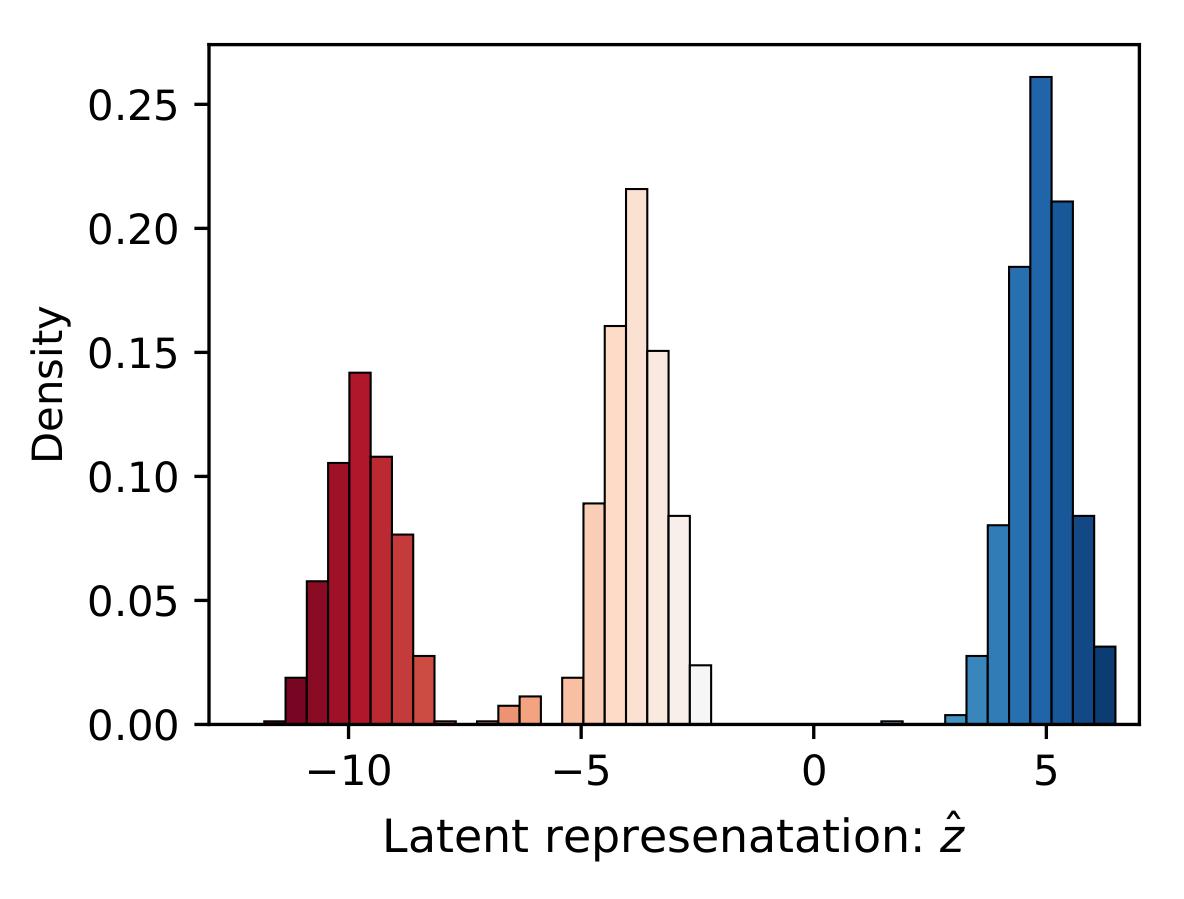}}
\end{subfigure}\hfill
\begin{subfigure}[Density of $\hat{z}$ from \ref{fig:ThreeMode_Blobs} in blue. Density of $\tilde{z}^*$ (red) belongs to E(x) from \ref{fig:Counterfactuals_CCHVAE} \label{fig:TwoMode_Blobs}.]{
\includegraphics[width=0.30\columnwidth]{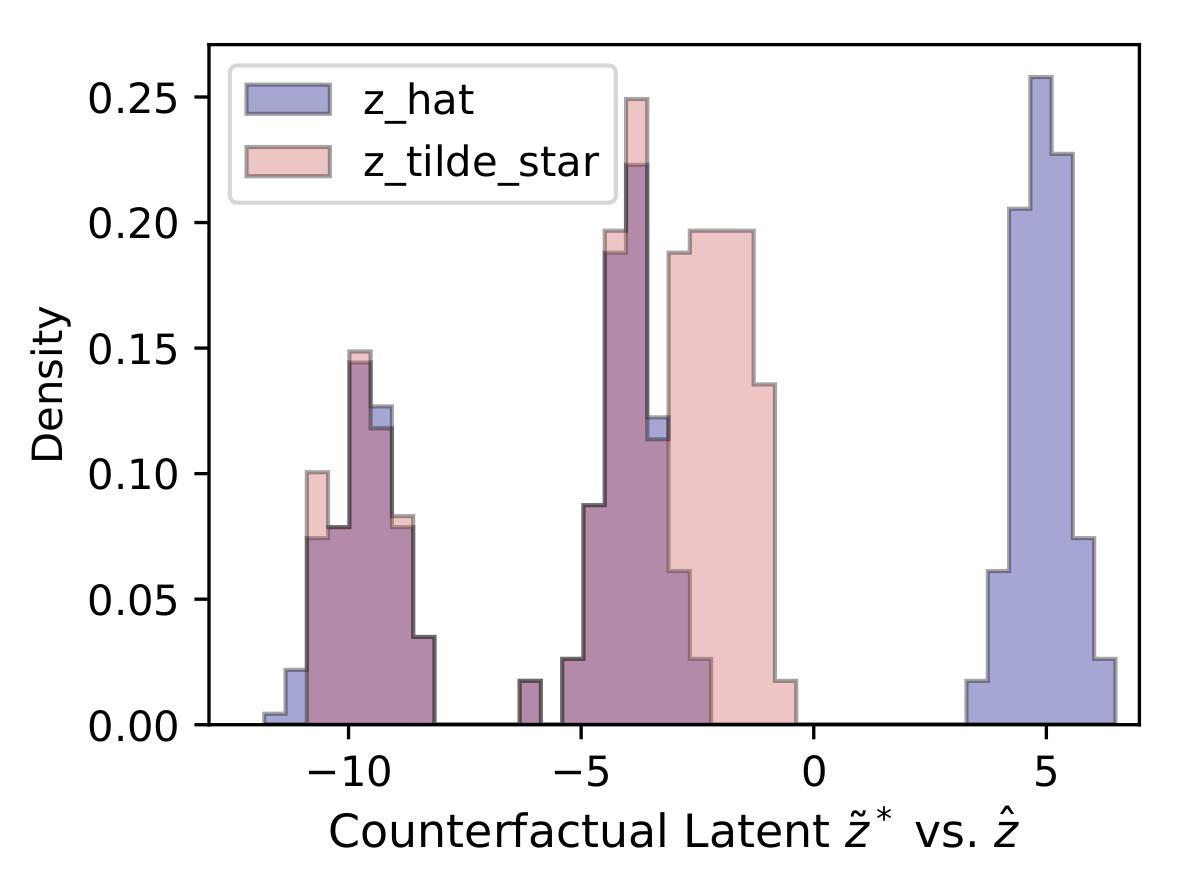}}
\end{subfigure}\vfill

\begin{subfigure}[True DGP. \label{fig:DGP_Blobs}]{
\includegraphics[width=0.30\columnwidth]{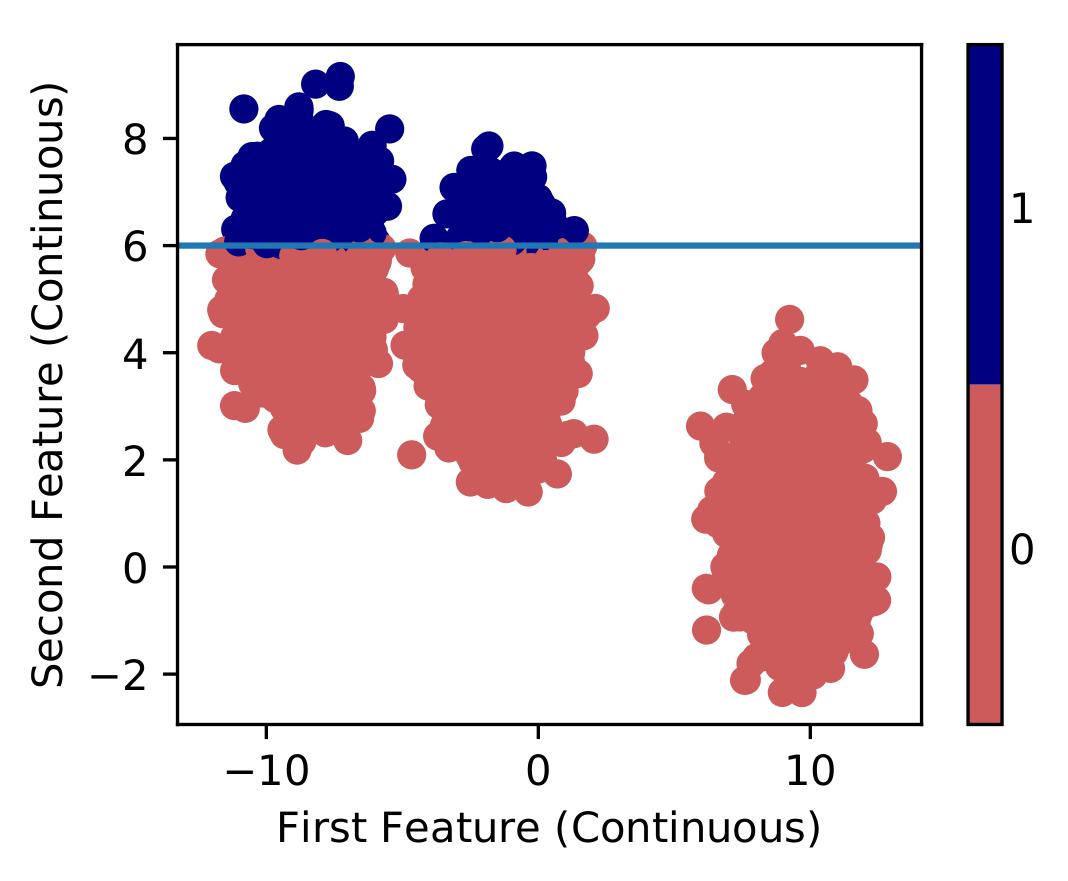}}
\end{subfigure}\hfill
\begin{subfigure}[Test data and {$E(\bm{x})$} by GS/AR (not shown). Upper right {$E(\bm{x})$} lie where no data is expected. \label{fig:Counterfactuals_Spangher}]{
\includegraphics[width=0.30\columnwidth]{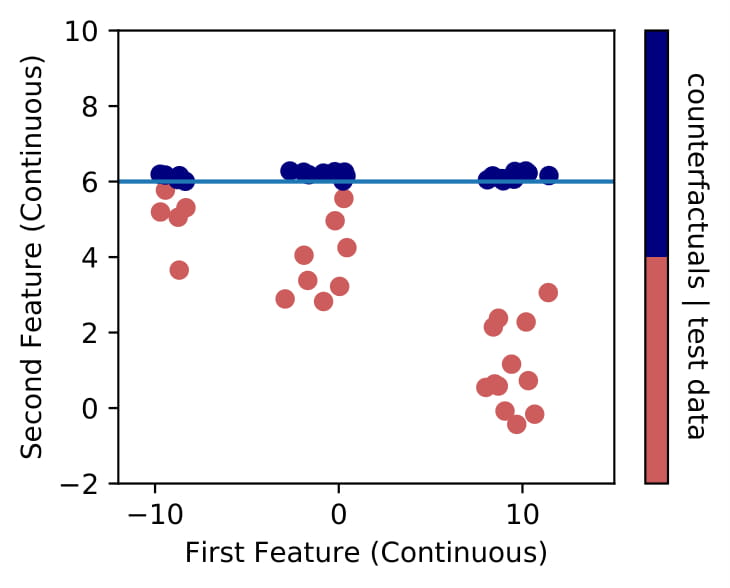}}
\end{subfigure}\hfill
\begin{subfigure}[Test data and {$E(\bm{x})$} by our cchvae. Most {$E(\bm{x})$} lie in high-density areas and are connected.\label{fig:Counterfactuals_CCHVAE}]{
\includegraphics[width=0.30\columnwidth]{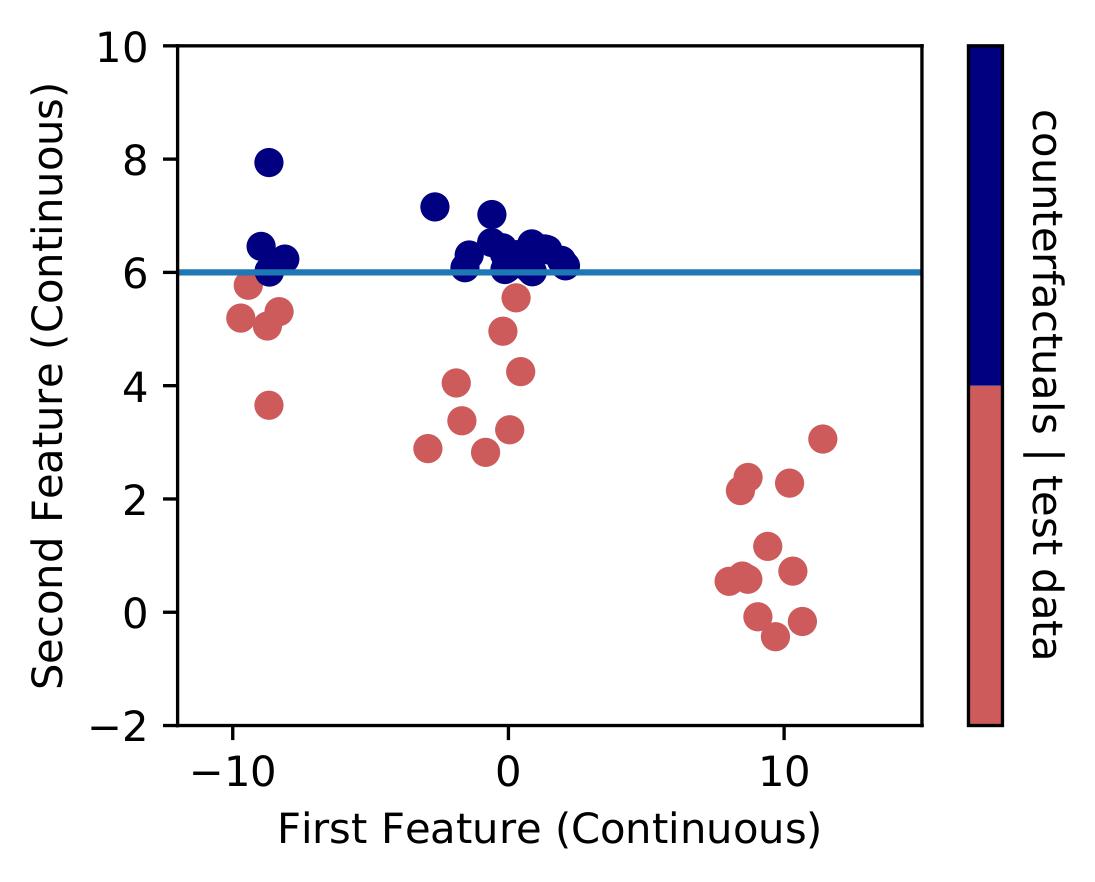}}
\end{subfigure}


\caption{Example \ref{ex:make_blobs}. Homogeneous features. Figure \ref{fig:TwoMode_Blobs} shows that generating close and meaningful counterfactuals amounts to finding the closest latent code from only 2 of the 3 modes of the latent distribution.}
\label{fig:ThreeBlobs}
\end{figure}

\paragraph{\textbf{Homogeneous features}.}
We begin by describing a data generating processes (DGP) for which it can be difficult to identify faithful counterfactuals. Example \ref{ex:make_blobs} corresponds to the case when all features are numerical. We generate 10000 observations from this DGP. We assume that the constant classifier $I(x_2>6)$ is given to us and our goal is to find counterfactuals for observations with 0-labels. Figure \ref{fig:Recon_Blobs} shows the reconstructed training data. The true DGP is shown in figure \ref{fig:DGP_Blobs}.

\begin{example}[\textbf{\emph{make blobs}}] We generate $\bm{x} = [x_1, x_2]$ from a mixture of 3 Gaussians with $\bm{\mu}=[\mu_1,\mu_2,\mu_3]$ and $\bm{\sigma} =  [\sigma_1,\sigma_2,\sigma_3] = [1,1,1]$ with a fixed seed. The response $y$ is generated from $Pr(y=1|X) = I(x_2 > 6)$, where $I(\cdot)$ denotes the indicator function.\label{ex:make_blobs}
\end{example}

Figure \ref{fig:Counterfactuals_Spangher} shows test data and their generated counterfactuals from AR and GS. For values from the lower right (blue) cluster in figure \ref{fig:Recon_Blobs}, both AR and GS suggest $E(\bm{x})$ that lie in the top right corner (figure \ref{fig:Counterfactuals_Spangher}). Since both GS and AR generate almost identical values, we report the results for GS only. AR and GS favour sparse $E(\bm{x})$, meaning they only require changes along the second feature axis. However, it is apparent that the upper right corner $E(\bm{x})$ are not attainable -- according to the DGP no data lives in this region. In contrast, our C-CHVAE suggests $E(\bm{x})$ that lie in regions of high data density, figure \ref{fig:Counterfactuals_CCHVAE}. 

To gain a better understanding of our method consider figure \ref{fig:ThreeMode_Blobs}. It shows the density of the estimated latent variable $\hat{z}$. The colours correspond to the clusters in the reconstructed data of figure \ref{fig:Recon_Blobs}. In figure \ref{fig:TwoMode_Blobs}, the counterfactual latent density $\tilde{z}^*$, i.e. the density of the latent variables from the counterfactuals $\tilde{x}^*$, is depicted on top of the density of $\hat{z}$. It shows that the density of $\tilde{z}^*$ is concentrated on the two modes which generate data that lies close to the decision boundary of the DGP. 




\subsection{Real world data sets}\label{sec:real_world}
For our real world experiments we choose 2 credit data sets; a processed version of the ``\texttt{Give me some credit}'' data set and the \emph{Home Equity Line of Credit} \texttt{(HELOC)} data set.\footnote{\url{https://www.kaggle.com/brycecf/give-me-some-credit-dataset.}}\footnote{\url{https://community.fico.com/s/explainable-machine-learning-challenge}.} For the former, the target variable records whether individuals experience financial distress within a period of two years, in the latter case one uses the applicants' information from credit reports to predict whether they will repay the \texttt{HELOC} account within a fixed time window. Both data sets are standard in the literature \citep{grath2018interpretable,spangher2018actionable,russell2019efficient} and are described in more detail in our github repository.

While GS works for different classifiers, the AR and HCLS algorithms do not. To also compare our results with AR we follow \citet{spangher2018actionable} and choose an $\ell_2$-penalized logistic regression model. For HCLS, we use SVM with a linear kernel when possible.

\paragraph{\textbf{``Give Me Some Credit''} (GMSC)} \label{sec:give_me_some}

\begin{figure}
\centering
\subfigure[Local outlier factor score \label{fig:lof_giveme}]{\includegraphics[width=0.45\columnwidth]{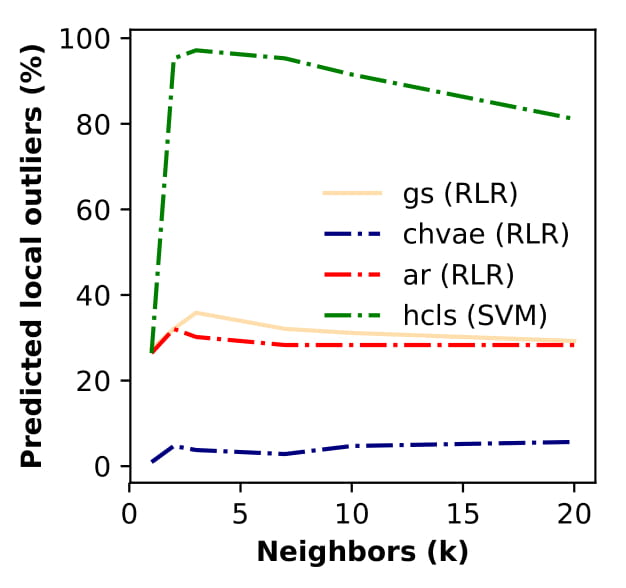}}%
\hfill
\subfigure[Connectedness score \label{fig:connected_giveme}]{\includegraphics[width=0.45\columnwidth]{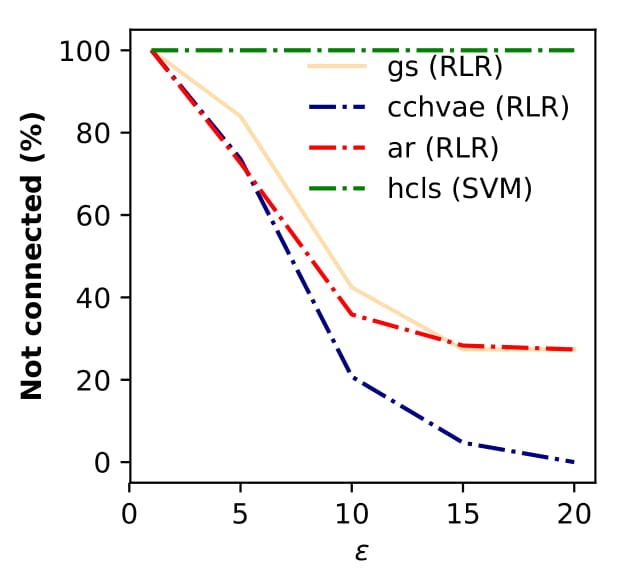}}%
\caption{\textbf{Faithfulnes relative to $\bm{x}_{} \in H^+ \cap D^+$} for \texttt{GMSC}.}
\label{fig:giveme_faithfulnes}

\subfigure[Local outlier factor score \label{fig:lof_heloc}]{\includegraphics[width=0.45\columnwidth]{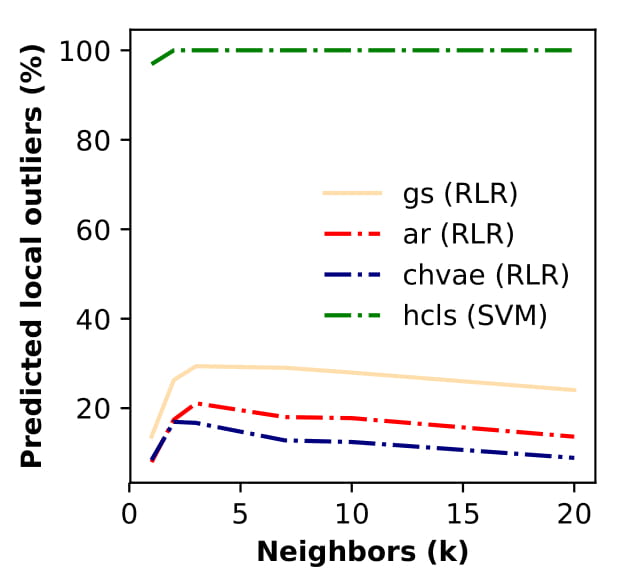}}%
\hfill
\subfigure[Connectedness score \label{fig:connected_heloc}]{\includegraphics[width=0.45\columnwidth]{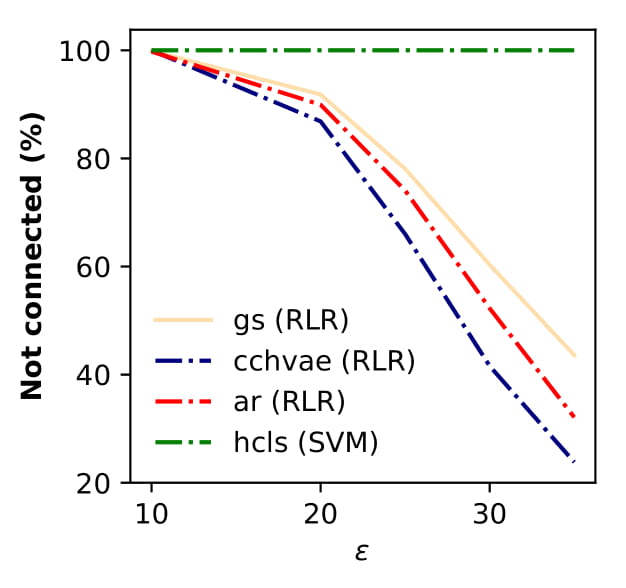}}%
\caption{\textbf{Faithfulness relative to $\bm{x}_{} \in H^+ \cap D^+$} for \texttt{HELOC} data.}
\label{fig:heloc_faithfulnes}
\end{figure}

\begin{figure}
\centering
\subfigure[AR \label{fig:total_ar}]{\includegraphics[width=0.38\columnwidth]{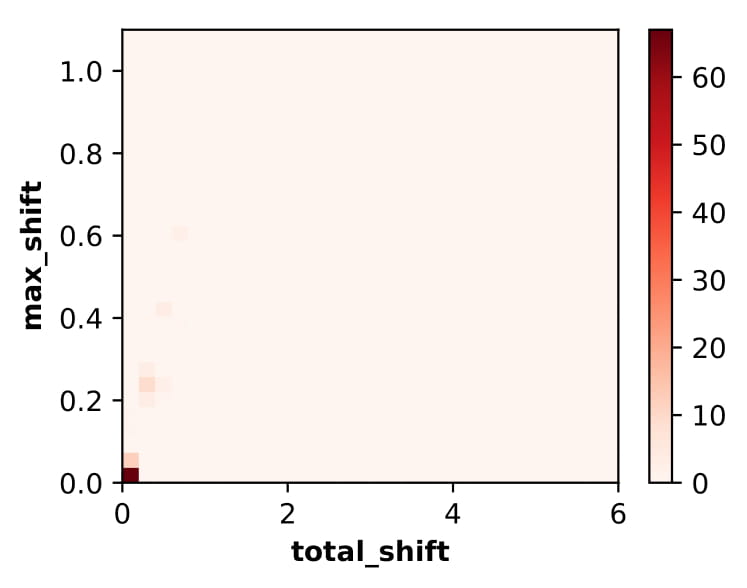}}%
\hfill
\subfigure[C-CHVAE \label{fig:total_cchvae}]{\includegraphics[width=0.38\columnwidth]{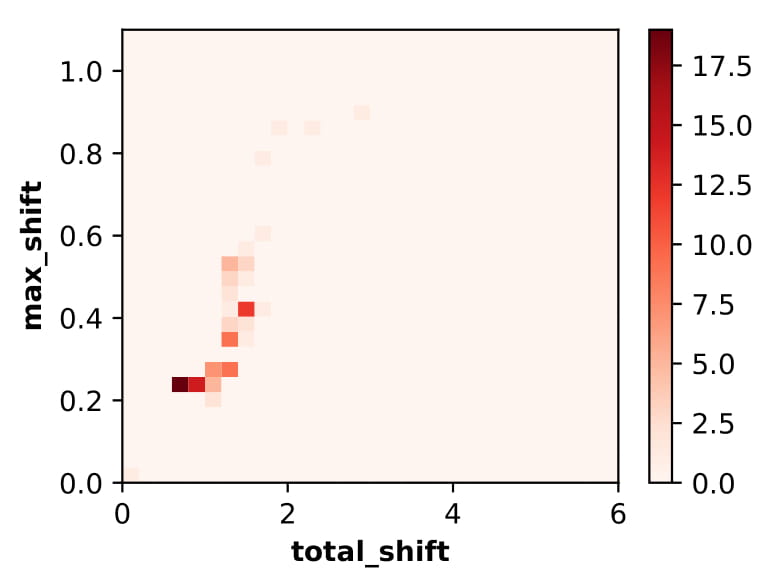}}%
\vfill
\subfigure[GS \label{fig:total_gs}]{\includegraphics[width=0.38\columnwidth]{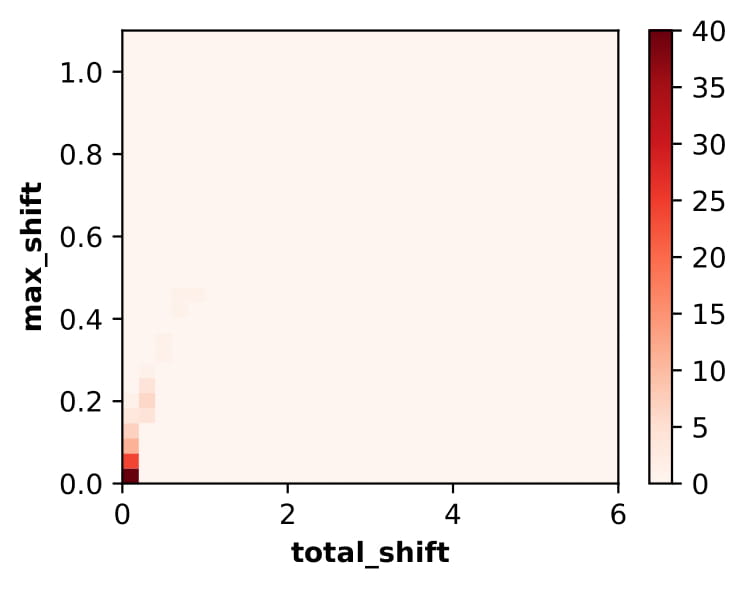}}%
\hfill 
\subfigure[HCLS \label{fig:total_hcls}]{\includegraphics[width=0.38\columnwidth]{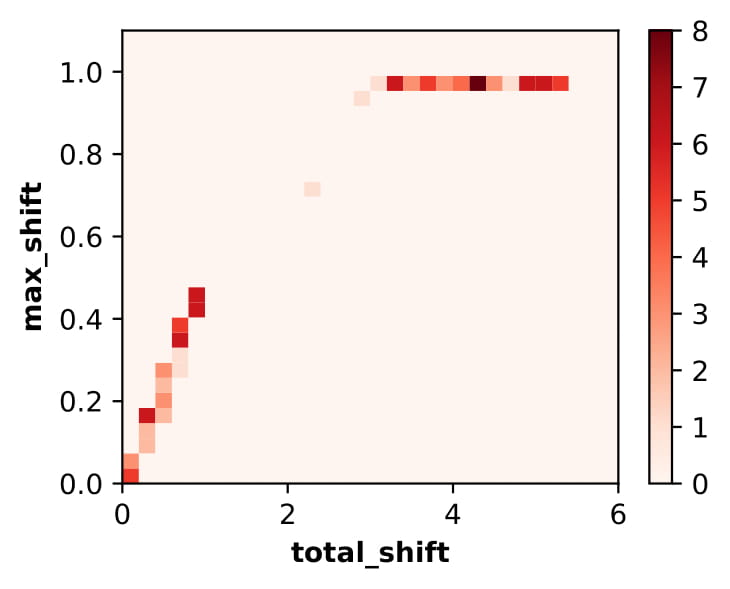}}%
\caption{\textbf{Total shift vs.\ max.\ shift} for $E(\bm{x})$ on \texttt{GMSC} data.}
\label{fig:giveme_2d_difficulty}

\subfigure[AR \label{fig:total_ar_heloc}]{\includegraphics[width=0.38\columnwidth]{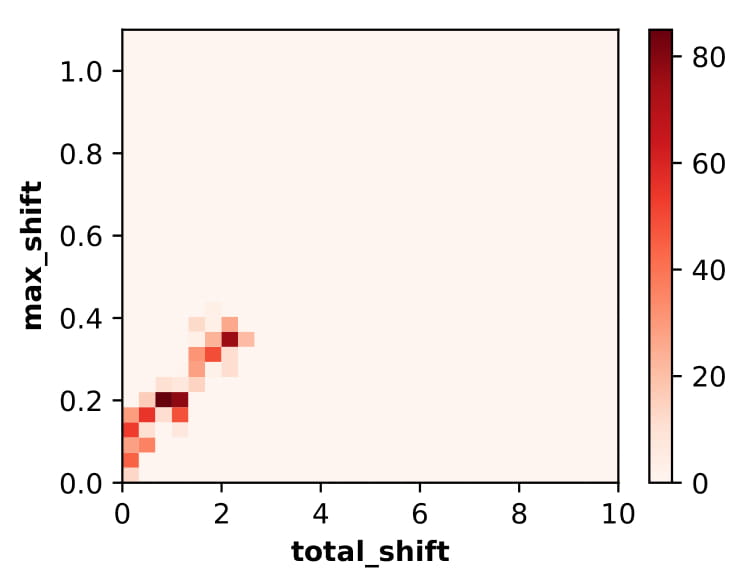}}%
\hfill
\subfigure[C-CHVAE \label{fig:total_cchvae_heloc}]{\includegraphics[width=0.38\columnwidth]{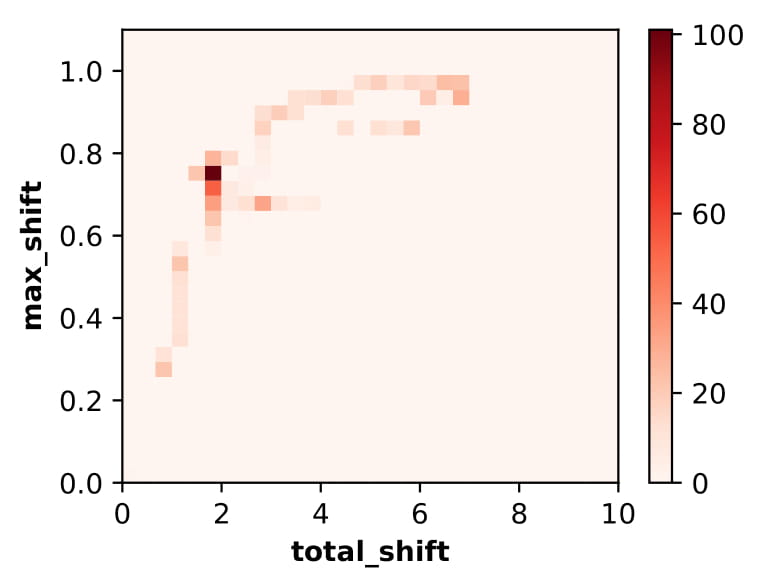}}%
\vfill
\subfigure[GS \label{fig:total_gs_heloc}]{\includegraphics[width=0.38\columnwidth]{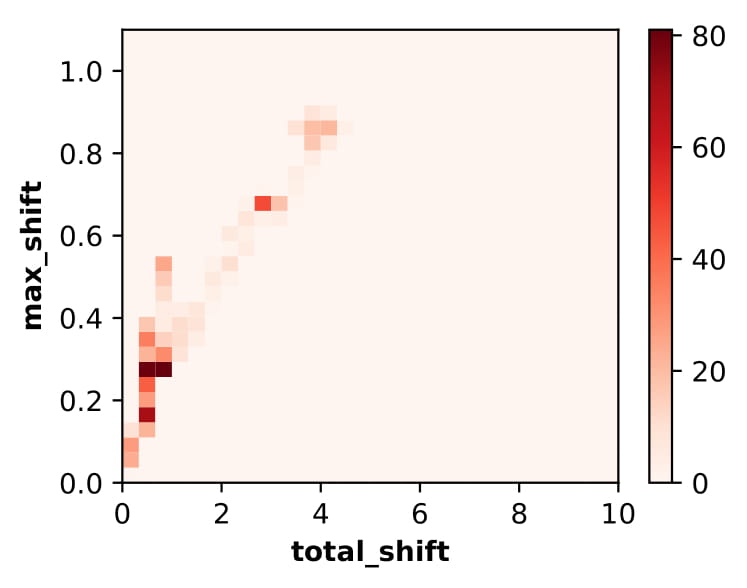}}%
\hfill 
\subfigure[HCLS  \label{fig:total_hcls_heloc}]{\includegraphics[width=0.38\columnwidth]{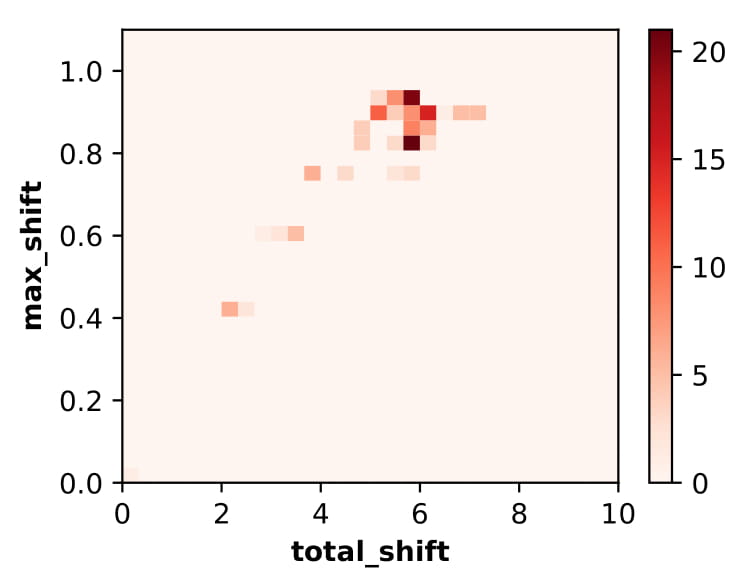}}%
\caption{\textbf{Total shift vs.\ max.\ shift} for $E(\bm{x})$ on \texttt{HELOC} data.}
\label{fig:heloc_2d_difficulty}
\end{figure}

For this data set, GS and AR produce very similar results in terms of faithfulness. HCLS performs worst and C-CHVAE (our's) outperforms all other methods. In terms of the local outlier score, the difference gets as high as 20 percentage points (figure \ref{fig:lof_giveme}). With respect to the connectedness score the difference grows larger for large $\epsilon$ (figure \ref{fig:connected_giveme}). In terms of \emph{difficulty}, it the C-CHVAE's faithfully generated counterfactuals come at the cost of greater TS and MS (figure \ref{fig:giveme_2d_difficulty}). 

\paragraph{\textbf{\texttt{HELOC}}}\label{sec:heloc}

With respect to \emph{counterfactual faithfulness}, the C-CHVAE outperforms all other methods for both measures and all parameter choices (figure \ref{fig:heloc_faithfulnes}). Again, HCLS is not performing well; one reason could lie in the fact that one needs to specify the directions in which all free features are allowed to change. This seems to require very careful choices. Moreover, it is likely to restrict the counterfactual suggestions, leading to counterfactuals that might look less typical, which is what \emph{faithfulness} measures. In terms of \emph{difficulty}, the pattern is similar to the one above (see figure \ref{fig:heloc_2d_difficulty}). The C-CHVAE tends to make suggestions with higher MS. This time, to obtain \emph{faithful counterfactuals} we are paying a price in terms of higher MS.

\section{Conclusion and future Work}\label{sec:conclusion}
We have introduced a general-purpose framework for generating counterfactuals; in particular, the fact that our method works for tabular data without the specification of distance or cost functions in the input space allows practitioners and researchers to adapt this work to a wide variety of applications.
To do so, several avenues for future work open up. First, all existing methods make recommendations of how features would need to be altered to receive a desired result, but none of these methods give associated input importance. And second, it would be desirable to formalize the tradeoff between the autoencoder capacity and counterfactual faithulness.

\begin{acks}
We would like to thank Lars Holdijk and Michael Lohaus for insightful discussions and Alfredo Nazabal for his assistance in running the HVAE.
\end{acks}


    

\bibliographystyle{ACM-Reference-Format}
\bibliography{sample-base}


\begin{thebibliography}{23}


\ifx \showCODEN    \undefined \def \showCODEN     #1{\unskip}     \fi
\ifx \showDOI      \undefined \def \showDOI       #1{#1}\fi
\ifx \showISBNx    \undefined \def \showISBNx     #1{\unskip}     \fi
\ifx \showISBNxiii \undefined \def \showISBNxiii  #1{\unskip}     \fi
\ifx \showISSN     \undefined \def \showISSN      #1{\unskip}     \fi
\ifx \showLCCN     \undefined \def \showLCCN      #1{\unskip}     \fi
\ifx \shownote     \undefined \def \shownote      #1{#1}          \fi
\ifx \showarticletitle \undefined \def \showarticletitle #1{#1}   \fi
\ifx \showURL      \undefined \def \showURL       {\relax}        \fi
\providecommand\bibfield[2]{#2}
\providecommand\bibinfo[2]{#2}
\providecommand\natexlab[1]{#1}
\providecommand\showeprint[2][]{arXiv:#2}

\bibitem[\protect\citeauthoryear{Agarwal, Beygelzimer, Dud{\'\i}k, Langford,
  and Wallach}{Agarwal et~al\mbox{.}}{2019}]%
        {agarwal2018reductions}
\bibfield{author}{\bibinfo{person}{Alekh Agarwal}, \bibinfo{person}{Alina
  Beygelzimer}, \bibinfo{person}{Miroslav Dud{\'\i}k}, \bibinfo{person}{John
  Langford}, {and} \bibinfo{person}{Hanna Wallach}.}
  \bibinfo{year}{2019}\natexlab{}.
\newblock \showarticletitle{A reductions approach to fair classification}. In
  \bibinfo{booktitle}{\emph{ICML}}.
\newblock


\bibitem[\protect\citeauthoryear{Akhtar and Mian}{Akhtar and Mian}{2018}]%
        {akhtar2018threat}
\bibfield{author}{\bibinfo{person}{Naveed Akhtar} {and} \bibinfo{person}{Ajmal
  Mian}.} \bibinfo{year}{2018}\natexlab{}.
\newblock \showarticletitle{Threat of adversarial attacks on deep learning in
  computer vision: A survey}.
\newblock \bibinfo{journal}{\emph{IEEE Access}}  \bibinfo{volume}{6}
  (\bibinfo{year}{2018}), \bibinfo{pages}{14410--14430}.
\newblock


\bibitem[\protect\citeauthoryear{Brown, Man{\'e}, Roy, Abadi, and Gilmer}{Brown
  et~al\mbox{.}}{2017}]%
        {brown2017adversarial}
\bibfield{author}{\bibinfo{person}{Tom~B Brown}, \bibinfo{person}{Dandelion
  Man{\'e}}, \bibinfo{person}{Aurko Roy}, \bibinfo{person}{Mart{\'\i}n Abadi},
  {and} \bibinfo{person}{Justin Gilmer}.} \bibinfo{year}{2017}\natexlab{}.
\newblock \showarticletitle{Adversarial patch}.
\newblock \bibinfo{journal}{\emph{arXiv preprint arXiv:1712.09665}}
  (\bibinfo{year}{2017}).
\newblock


\bibitem[\protect\citeauthoryear{Grath, Costabello, Van, Sweeney, Kamiab, Shen,
  and Lecue}{Grath et~al\mbox{.}}{2018}]%
        {grath2018interpretable}
\bibfield{author}{\bibinfo{person}{Rory~Mc Grath}, \bibinfo{person}{Luca
  Costabello}, \bibinfo{person}{Chan~Le Van}, \bibinfo{person}{Paul Sweeney},
  \bibinfo{person}{Farbod Kamiab}, \bibinfo{person}{Zhao Shen}, {and}
  \bibinfo{person}{Freddy Lecue}.} \bibinfo{year}{2018}\natexlab{}.
\newblock \showarticletitle{Interpretable Credit Application Predictions With
  Counterfactual Explanations}.
\newblock \bibinfo{journal}{\emph{NeurIPS workshop: Challenges and
  Opportunities for AI in Financial Services: the Impact of Fairness,
  Explainability, Accuracy, and Privacy}} (\bibinfo{year}{2018}).
\newblock


\bibitem[\protect\citeauthoryear{Grgic-Hlaca, Redmiles, Gummadi, and
  Weller}{Grgic-Hlaca et~al\mbox{.}}{2018}]%
        {grgic2018human}
\bibfield{author}{\bibinfo{person}{Nina Grgic-Hlaca}, \bibinfo{person}{Elissa~M
  Redmiles}, \bibinfo{person}{Krishna~P Gummadi}, {and} \bibinfo{person}{Adrian
  Weller}.} \bibinfo{year}{2018}\natexlab{}.
\newblock \showarticletitle{Human perceptions of fairness in algorithmic
  decision making: A case study of criminal risk prediction}. In
  \bibinfo{booktitle}{\emph{Proceedings of the 2018 World Wide Web
  Conference}}. International World Wide Web Conferences Steering Committee,
  \bibinfo{pages}{903--912}.
\newblock


\bibitem[\protect\citeauthoryear{Harman and Lacko}{Harman and Lacko}{2010}]%
        {harman2010decompositional}
\bibfield{author}{\bibinfo{person}{Radoslav Harman} {and}
  \bibinfo{person}{Vladim{\'\i}r Lacko}.} \bibinfo{year}{2010}\natexlab{}.
\newblock \showarticletitle{On decompositional algorithms for uniform sampling
  from n-spheres and n-balls}.
\newblock \bibinfo{journal}{\emph{Journal of Multivariate Analysis}}
  \bibinfo{volume}{101}, \bibinfo{number}{10} (\bibinfo{year}{2010}),
  \bibinfo{pages}{2297--2304}.
\newblock


\bibitem[\protect\citeauthoryear{Ivanov, Figurnov, and Vetrov}{Ivanov
  et~al\mbox{.}}{2018}]%
        {ivanov2018variational}
\bibfield{author}{\bibinfo{person}{Oleg Ivanov}, \bibinfo{person}{Michael
  Figurnov}, {and} \bibinfo{person}{Dmitry Vetrov}.}
  \bibinfo{year}{2018}\natexlab{}.
\newblock \showarticletitle{Variational Autoencoder with Arbitrary
  Conditioning}.
\newblock \bibinfo{journal}{\emph{arXiv preprint arXiv:1806.02382}}
  (\bibinfo{year}{2018}).
\newblock


\bibitem[\protect\citeauthoryear{Joshi, Koyejo, Vijitbenjaronk, Kim, and
  Ghosh}{Joshi et~al\mbox{.}}{2019}]%
        {joshi2019towards}
\bibfield{author}{\bibinfo{person}{Shalmali Joshi}, \bibinfo{person}{Oluwasanmi
  Koyejo}, \bibinfo{person}{Warut Vijitbenjaronk}, \bibinfo{person}{Been Kim},
  {and} \bibinfo{person}{Joydeep Ghosh}.} \bibinfo{year}{2019}\natexlab{}.
\newblock \showarticletitle{Towards Realistic Individual Recourse and
  Actionable Explanations in Black-Box Decision Making Systems}.
\newblock \bibinfo{journal}{\emph{arXiv preprint arXiv:1907.09615}}
  (\bibinfo{year}{2019}).
\newblock


\bibitem[\protect\citeauthoryear{Kingma and Welling}{Kingma and
  Welling}{2013}]%
        {kingma2013auto}
\bibfield{author}{\bibinfo{person}{Diederik~P Kingma} {and}
  \bibinfo{person}{Max Welling}.} \bibinfo{year}{2013}\natexlab{}.
\newblock \showarticletitle{Auto-encoding variational bayes}.
\newblock \bibinfo{journal}{\emph{Proceedings of the 2nd International
  Conference on Learning Representations (ICLR)}} (\bibinfo{year}{2013}).
\newblock


\bibitem[\protect\citeauthoryear{Lash, Lin, Street, Robinson, and Ohlmann}{Lash
  et~al\mbox{.}}{2017}]%
        {lash2017generalized}
\bibfield{author}{\bibinfo{person}{Michael~T Lash}, \bibinfo{person}{Qihang
  Lin}, \bibinfo{person}{Nick Street}, \bibinfo{person}{Jennifer~G Robinson},
  {and} \bibinfo{person}{Jeffrey Ohlmann}.} \bibinfo{year}{2017}\natexlab{}.
\newblock \showarticletitle{Generalized inverse classification}. In
  \bibinfo{booktitle}{\emph{Proceedings of the 2017 SIAM International
  Conference on Data Mining}}. SIAM, \bibinfo{pages}{162--170}.
\newblock


\bibitem[\protect\citeauthoryear{Laugel, Lesot, Marsala, and Detyniecki}{Laugel
  et~al\mbox{.}}{2019}]%
        {laugel2019issues}
\bibfield{author}{\bibinfo{person}{Thibault Laugel},
  \bibinfo{person}{Marie-Jeanne Lesot}, \bibinfo{person}{Christophe Marsala},
  {and} \bibinfo{person}{Marcin Detyniecki}.} \bibinfo{year}{2019}\natexlab{}.
\newblock \showarticletitle{Issues with post-hoc counterfactual explanations: a
  discussion}.
\newblock \bibinfo{journal}{\emph{ICML Workshop on Human in the Loop Learning}}
  (\bibinfo{year}{2019}).
\newblock


\bibitem[\protect\citeauthoryear{Laugel, Lesot, Marsala, Renard, and
  Detyniecki}{Laugel et~al\mbox{.}}{2017}]%
        {laugel2017inverse}
\bibfield{author}{\bibinfo{person}{Thibault Laugel},
  \bibinfo{person}{Marie-Jeanne Lesot}, \bibinfo{person}{Christophe Marsala},
  \bibinfo{person}{Xavier Renard}, {and} \bibinfo{person}{Marcin Detyniecki}.}
  \bibinfo{year}{2017}\natexlab{}.
\newblock \showarticletitle{Inverse Classification for Comparison-based
  Interpretability in Machine Learning}.
\newblock \bibinfo{journal}{\emph{arXiv preprint arXiv:1712.08443}}
  (\bibinfo{year}{2017}).
\newblock


\bibitem[\protect\citeauthoryear{Makhzani, Shlens, Jaitly, Goodfellow, and
  Frey}{Makhzani et~al\mbox{.}}{2015}]%
        {makhzani2015adversarial}
\bibfield{author}{\bibinfo{person}{Alireza Makhzani}, \bibinfo{person}{Jonathon
  Shlens}, \bibinfo{person}{Navdeep Jaitly}, \bibinfo{person}{Ian Goodfellow},
  {and} \bibinfo{person}{Brendan Frey}.} \bibinfo{year}{2015}\natexlab{}.
\newblock \showarticletitle{Adversarial autoencoders}.
\newblock \bibinfo{journal}{\emph{arXiv preprint arXiv:1511.05644}}
  (\bibinfo{year}{2015}).
\newblock


\bibitem[\protect\citeauthoryear{Nazabal, Olmos, Ghahramani, and
  Valera}{Nazabal et~al\mbox{.}}{2018}]%
        {nazabal2018handling}
\bibfield{author}{\bibinfo{person}{Alfredo Nazabal}, \bibinfo{person}{Pablo~M
  Olmos}, \bibinfo{person}{Zoubin Ghahramani}, {and} \bibinfo{person}{Isabel
  Valera}.} \bibinfo{year}{2018}\natexlab{}.
\newblock \showarticletitle{Handling incomplete heterogeneous data using VAEs}.
\newblock \bibinfo{journal}{\emph{arXiv preprint arXiv:1807.03653}}
  (\bibinfo{year}{2018}).
\newblock


\bibitem[\protect\citeauthoryear{Pawelczyk, Haug, Broelemann, and
  Kasneci}{Pawelczyk et~al\mbox{.}}{2019}]%
        {pawelczyk2019user}
\bibfield{author}{\bibinfo{person}{Martin Pawelczyk}, \bibinfo{person}{Johannes
  Haug}, \bibinfo{person}{Klaus Broelemann}, {and} \bibinfo{person}{Gjergji
  Kasneci}.} \bibinfo{year}{2019}\natexlab{}.
\newblock \showarticletitle{Towards User Empowerment}.
\newblock \bibinfo{journal}{\emph{NeurIPS Workshop on Human-Centric Machine
  Learning}} (\bibinfo{year}{2019}).
\newblock


\bibitem[\protect\citeauthoryear{Regitz-Zagrosek}{Regitz-Zagrosek}{2012}]%
        {regitz2012sex}
\bibfield{author}{\bibinfo{person}{Vera Regitz-Zagrosek}.}
  \bibinfo{year}{2012}\natexlab{}.
\newblock \showarticletitle{Sex and gender differences in health}.
\newblock \bibinfo{journal}{\emph{EMBO reports}} \bibinfo{volume}{13},
  \bibinfo{number}{7} (\bibinfo{year}{2012}), \bibinfo{pages}{596--603}.
\newblock


\bibitem[\protect\citeauthoryear{Russell}{Russell}{2019}]%
        {russell2019efficient}
\bibfield{author}{\bibinfo{person}{Christopher Russell}.}
  \bibinfo{year}{2019}\natexlab{}.
\newblock \showarticletitle{Efficient Search for Diverse Coherent
  Explanations}. In \bibinfo{booktitle}{\emph{Proceedings of the Conference on
  Fairness, Accountability, and Transparency}}. ACM FAT,
  \bibinfo{pages}{20--28}.
\newblock


\bibitem[\protect\citeauthoryear{Sohn, Lee, and Yan}{Sohn
  et~al\mbox{.}}{2015}]%
        {sohn2015learning}
\bibfield{author}{\bibinfo{person}{Kihyuk Sohn}, \bibinfo{person}{Honglak Lee},
  {and} \bibinfo{person}{Xinchen Yan}.} \bibinfo{year}{2015}\natexlab{}.
\newblock \showarticletitle{Learning structured output representation using
  deep conditional generative models}. In \bibinfo{booktitle}{\emph{Advances in
  neural information processing systems}}. \bibinfo{pages}{3483--3491}.
\newblock


\bibitem[\protect\citeauthoryear{Tolomei, Silvestri, Haines, and
  Lalmas}{Tolomei et~al\mbox{.}}{2017}]%
        {tolomei2017interpretable}
\bibfield{author}{\bibinfo{person}{Gabriele Tolomei}, \bibinfo{person}{Fabrizio
  Silvestri}, \bibinfo{person}{Andrew Haines}, {and} \bibinfo{person}{Mounia
  Lalmas}.} \bibinfo{year}{2017}\natexlab{}.
\newblock \showarticletitle{Interpretable predictions of tree-based ensembles
  via actionable feature tweaking}. In \bibinfo{booktitle}{\emph{Proceedings of
  the 23rd ACM SIGKDD international conference on knowledge discovery and data
  mining}}. ACM, \bibinfo{pages}{465--474}.
\newblock


\bibitem[\protect\citeauthoryear{Tolstikhin, Bousquet, Gelly, and
  Schoelkopf}{Tolstikhin et~al\mbox{.}}{2017}]%
        {tolstikhin2017wasserstein}
\bibfield{author}{\bibinfo{person}{Ilya Tolstikhin}, \bibinfo{person}{Olivier
  Bousquet}, \bibinfo{person}{Sylvain Gelly}, {and} \bibinfo{person}{Bernhard
  Schoelkopf}.} \bibinfo{year}{2017}\natexlab{}.
\newblock \showarticletitle{Wasserstein auto-encoders}.
\newblock \bibinfo{journal}{\emph{arXiv preprint arXiv:1711.01558}}
  (\bibinfo{year}{2017}).
\newblock


\bibitem[\protect\citeauthoryear{Ustun, Spangher, and Liu}{Ustun
  et~al\mbox{.}}{2019}]%
        {spangher2018actionable}
\bibfield{author}{\bibinfo{person}{Berk Ustun}, \bibinfo{person}{Alexander
  Spangher}, {and} \bibinfo{person}{Yang Liu}.}
  \bibinfo{year}{2019}\natexlab{}.
\newblock \showarticletitle{Actionable recourse in linear classification}. In
  \bibinfo{booktitle}{\emph{Proceedings of the Conference on Fairness,
  Accountability, and Transparency}}. ACM, \bibinfo{pages}{10--19}.
\newblock


\bibitem[\protect\citeauthoryear{Wachter, Mittelstadt, and Russell}{Wachter
  et~al\mbox{.}}{2017}]%
        {wachter2017counterfactual}
\bibfield{author}{\bibinfo{person}{Sandra Wachter}, \bibinfo{person}{Brent
  Mittelstadt}, {and} \bibinfo{person}{Chris Russell}.}
  \bibinfo{year}{2017}\natexlab{}.
\newblock \showarticletitle{Counterfactual explanations without opening the
  black box: automated decisions and the GDPR}.
\newblock \bibinfo{journal}{\emph{Harvard Journal of Law \& Technology}}
  \bibinfo{volume}{31}, \bibinfo{number}{2} (\bibinfo{year}{2017}),
  \bibinfo{pages}{2018}.
\newblock


\bibitem[\protect\citeauthoryear{Zafar, Valera, Gomez~Rodriguez, and
  Gummadi}{Zafar et~al\mbox{.}}{2017}]%
        {zafar2017fairness}
\bibfield{author}{\bibinfo{person}{Muhammad~Bilal Zafar},
  \bibinfo{person}{Isabel Valera}, \bibinfo{person}{Manuel Gomez~Rodriguez},
  {and} \bibinfo{person}{Krishna~P Gummadi}.} \bibinfo{year}{2017}\natexlab{}.
\newblock \showarticletitle{Fairness beyond disparate treatment \& disparate
  impact: Learning classification without disparate mistreatment}. In
  \bibinfo{booktitle}{\emph{Proceedings of the 26th International Conference on
  World Wide Web}}. International World Wide Web Conferences Steering
  Committee, \bibinfo{pages}{1171--1180}.
\newblock


\end{thebibliography}

\clearpage
\appendix
 \section{Counterfactual Search -- Algorithms} \label{sec:algorithm}
\subsection{Counterfactual search algorithm} \label{sec:algorithm}
As inputs, the algorithm requires any pretrained classifier $f$ and the trained decoder and encoder from the CHVAE. It returns the closest counterfactual. We note algorithm \ref{alg:StochasticSearch} uses a standard procedure to generate random numbers distributed uniformly over a sphere. \citet{laugel2017inverse} use a similar algorithm, but relative to their work, we look for the smallest change in the \emph{latent representation} $\bm{z}$ (not in input space) that would lead to a change in the predicted label. Thus, we sample observations $\tilde{\bm{z}}$ in $l_p$-spheres around the point $\hat{\bm{z}}$ until we find a counterfactual $\tilde{\bm{x}}^*$. For positive numbers $r_1$ and $r_2$, we define a $(r_1,r_2)$-sphere around $\hat{\bm{z}}$:
\begin{equation}
    S(\hat{\bm{z}}, r_1, r_2) = \{ \tilde{\bm{z}} \in \mathcal{Z}: r_1 \leq \lVert \hat{{z}}-\tilde{\bm{z}} \rVert \leq r_2 \}.
    \label{eq:sphere}
\end{equation}

In order to generate uniform random numbers over a sphere, we also use the
YPHL algorithm \citep{harman2010decompositional}. Their algorithm allows us to generate observations uniformly distributed over the unit-sphere. Next, one draws observations uniformly from $U[r_1,r_2]$, which are in turn used to rescale the distance between uniform sphere values and $\hat{\bm{z}}$. Eventually, we arrive at observations $\tilde{\bm{z}}$ that are uniformly distributed over $S(\hat{\bm{z}}, r_1, r_2)$.

Algorithm~\ref{alg:StochasticSearch} shows a counterfactual search procedure when the latent variable has a dense distribution. It is straightforward to adjust the algorithm to scenarios when one desires to generate multiple counterfactual examples, also known as \emph{flip sets} \citep{spangher2018actionable,russell2019efficient}. The idea is that the user can choose one counterfactual from a menu of different counterfactuals, which fits her preferences best.  

\begin{algorithm}[htb!]
  \caption{Stochastic Counterfactual Search For Latent Space}
  \label{alg:StochasticSearch}
\begin{algorithmic}
  \STATE {\bfseries Input:} $X_{train}$: training data; $x_{i,test}$: test observation; $f$: classifier trained on $X_{train}$; $m_{\hat{\theta}}$, $g_{\hat{\phi}}$: CHVAE encoder and decoder trained on $X_{train}$; $S$: number search samples; $\Delta r$: search radius.
  \STATE {\bfseries Initialize:} $f(x_{i,test})= \hat{y}_{i,test}$; $m_{\hat{\theta}}(x_{i,test})=\hat{z}_{i,test}$; $r = 0$; $\mathcal{C} = \varnothing; \hat{z}_{train, min} = \min_i \hat{z}_{i,train}; \hat{z}_{train, max} = \max_i \hat{z}_{i,train}$.
  \WHILE{$\mathcal{C} = \varnothing ~ \wedge \tilde{z}_{i,test} \in [\hat{z}_{train, min}, \hat{z}_{train,max}]$}
  \FOR{$j=1$ {\bfseries to} $J$}
    \item sample $\tilde{z}_{i,test}^j$ from $S(\hat{\bm{z}}_{i,test},r,\Delta r)$ in \eqref{eq:sphere} \COMMENT{Perturbed representation}
    
  \item $\tilde{x}^j_{i,test} = g_{\hat{\phi}}(\tilde{z}^j_{i,test})$ \COMMENT{Potential counterfactual}
  \item $\tilde{y}^j_{i,test} = f(\tilde{x}^j_{i,test})$
  \IF{$\tilde{y}^j_{i,test} \neq \hat{y}_{i,test}$}
  \STATE $\mathcal{C} \xleftarrow{} (\tilde{z}^j_{i,test}, \tilde{x}^j_{i, test}, \tilde{y}^j_{i, test})$
  \ENDIF
  \ENDFOR
  \IF{$\mathcal{C} = \varnothing$}
  \item $r = r + \Delta r$ \COMMENT{Push search range outward}
  \ELSIF{$\tilde{z}_{i,test} \not \in [\hat{z}_{train, min}, \hat{z}_{train,max}]$}
  \item {\bfseries Return:} \COMMENT{No counterfactual consistent with data distribution}
  \item $\mathcal{C} = \varnothing$
  \ELSE
  \item {\bfseries Return:} \COMMENT{Find 'closest' counterfactual}
  \item $\tilde{z}^*_{i,test} = \text{argmin}_{\tilde{z}_{test} \in \mathcal{C}} ||\tilde{z}_{test} - \hat{z}_{test}||$
  \item $\tilde{x}^*_{i,test} = g_{\hat{\phi}}(\tilde{z}_{i,test}^*)$, $ \tilde{y}_{i,test}^* = f(\tilde{x}_{i,test}^*)$
  \ENDIF
  \ENDWHILE
\end{algorithmic}
\end{algorithm}

\section{Common likelihood models}\label{sec:appendix_likelihood}

For the sake of completeness we enumerate a list of commonly used likelihood models for numerical and nominal features \citep{nazabal2018handling}:
\begin{itemize}
    \item \textbf{Real-valued data}. For real valued data, one usually assumes a Gaussian likelihood model such as,
    \begin{equation*}
        p(\bm{x}_{i,d}|\bm{\gamma}_{i,d}) = \mathcal{N}(\mu_d(\bm{z}_i), \sigma^2_d(\bm{z}_i)),
        \label{eq:lik_gauss}
    \end{equation*}
    where $\bm{\gamma}_{i,d} = \{\mu_d(\bm{z}_i), \sigma^2_d(\bm{z}_i)\}$ are modelled by the outputs of a DNN with inputs $\bm{z}$.
    
        \item \textbf{Positive real-valued data}. For positive real valued data, one can assume a log normal likelihood model such as,
    \begin{equation*}
        p(\bm{x}_{i,d}|\bm{\gamma}_{i,d}) = \log \mathcal{N}(\mu_d(\bm{z}_i), \sigma^2_d(\bm{z}_i)),
        \label{eq:lik_log_gauss}
    \end{equation*}
    where $\bm{\gamma}_{i,d} = \{\mu_d(\bm{z}_i), \sigma^2_d(\bm{z}_i)\}$.

            \item \textbf{Count data}. For count data, one can assume a Poisson likelihood model such as,
    \begin{equation*}
        p(\bm{x}_{i,d}|\bm{\gamma}_{i,d}) = Poisson(\lambda_d(\bm{z}_i)),
        \label{eq:lik_pois}
    \end{equation*}
    where $\bm{\gamma}_{i,d} = \{\lambda_d(\bm{z}_i)\}$.
                \item \textbf{Ordinal data}. For ordinal valued data, we use the same procedure as in \citep{nazabal2018handling}. 
    
    \item \textbf{Categorical data.} For categorical data one can assume a multinomial logit model, where the probability of every category $r$ is given by
    \begin{equation*}
     p(\bm{x}_{i,d} = r|\bm{\gamma}_{i,d}) = \frac{\exp^{(h_{d_r}(\bm{z}_i))}}{\sum_{r=1}^R \exp^{(h_{d_r}(\bm{z}_i))}},
        \label{eq:lik_cat}
    \end{equation*}
    with parameters $\bm{\gamma}_{i,d} = \{h_{d_0}(\bm{z}_i), h_{d_1}(\bm{z}_i), ..., h_{d_{R-1}}(\bm{z}_i) \}$ and $h_{d_0}(\bm{z}_i)=0$ to ensure identifiablity.

\end{itemize}

\section{Synthetic example}\label{sec:synthetic_examples}

\begin{example}[\textbf{\emph{discretized make moons}}] We generate the upper have circle $[x_1, x_2]$-pairs by $cos(i) \text{ for } i \in (0,\pi)$, which we round to the closest decimal. The lower have circle $[x_1, x_2]$-pairs are then generated by 1 - $sin(i) \text{ for } i \in (0,\pi)$, where we round $sin(i)$ to the closest integer. Both the upper and the lower half contain half of the observations each and $x_2$ is treated as categorical with 19 categories. The response $y$ is then generated from $Pr(y=1|X) = I(x_1 > 0)$.  \label{ex:make_moons} \end{example}

\begin{figure}
\centering
\subfigure[2-d histogram of data generating process \label{fig:dgp_2dhist_discrete}]{\includegraphics[width=0.30\columnwidth]{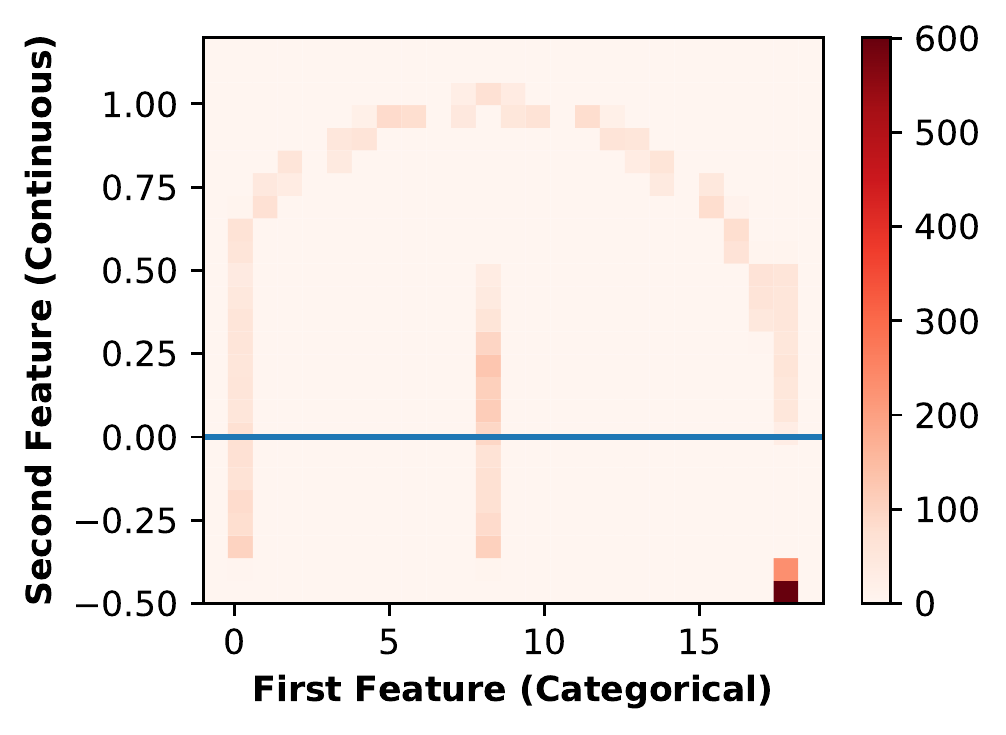}}
\hfill
\subfigure[Test data 2d-histogram from 0-class.\label{fig:dgp_2dhist_test}]{\includegraphics[width=0.30\columnwidth]{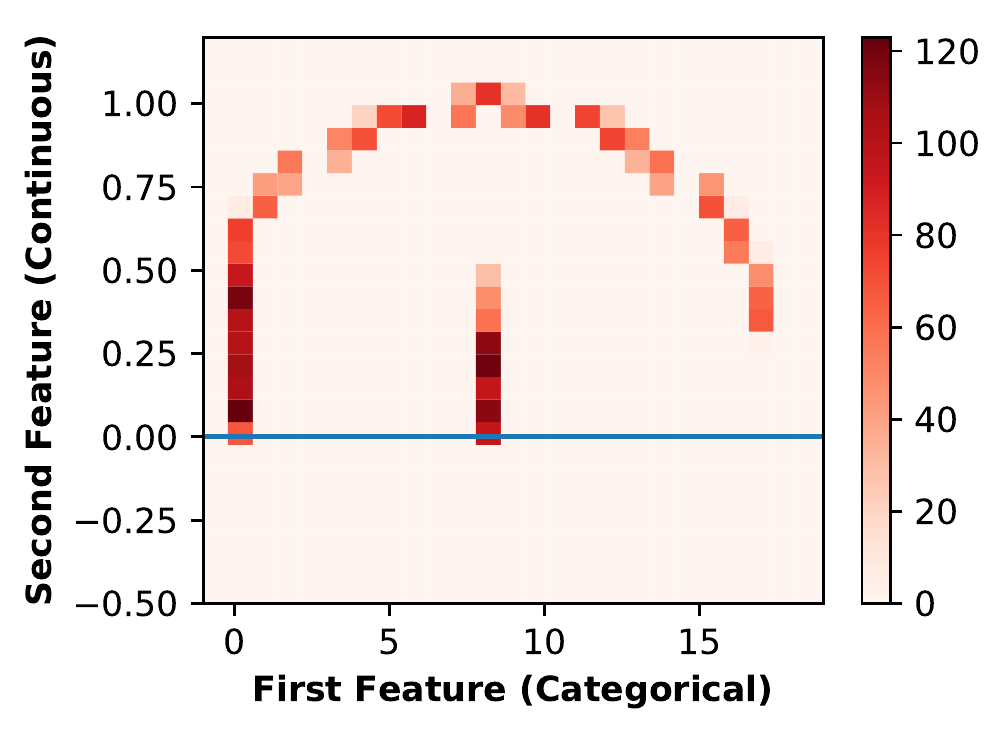}}%
\hfill
\subfigure[Reconstructed training data. \label{fig:discrete_recon}]{\includegraphics[width=0.30\columnwidth]{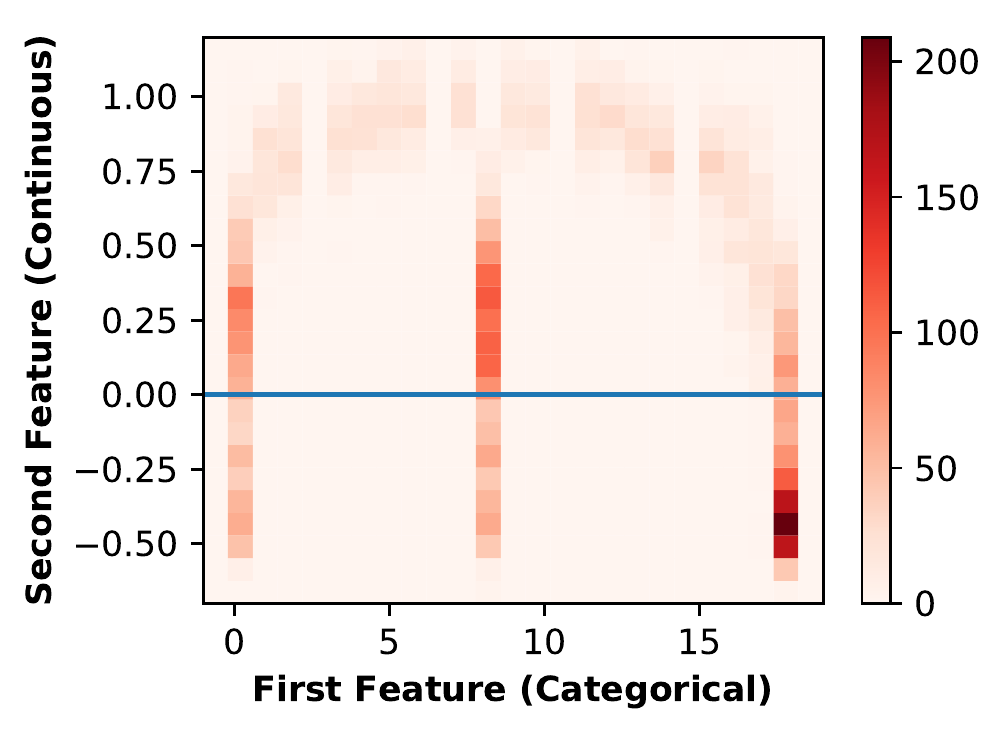}} %

\subfigure[$E(\bm{x})$ by ar. \label{fig:discrete_ar}]{\includegraphics[width=0.30\columnwidth]{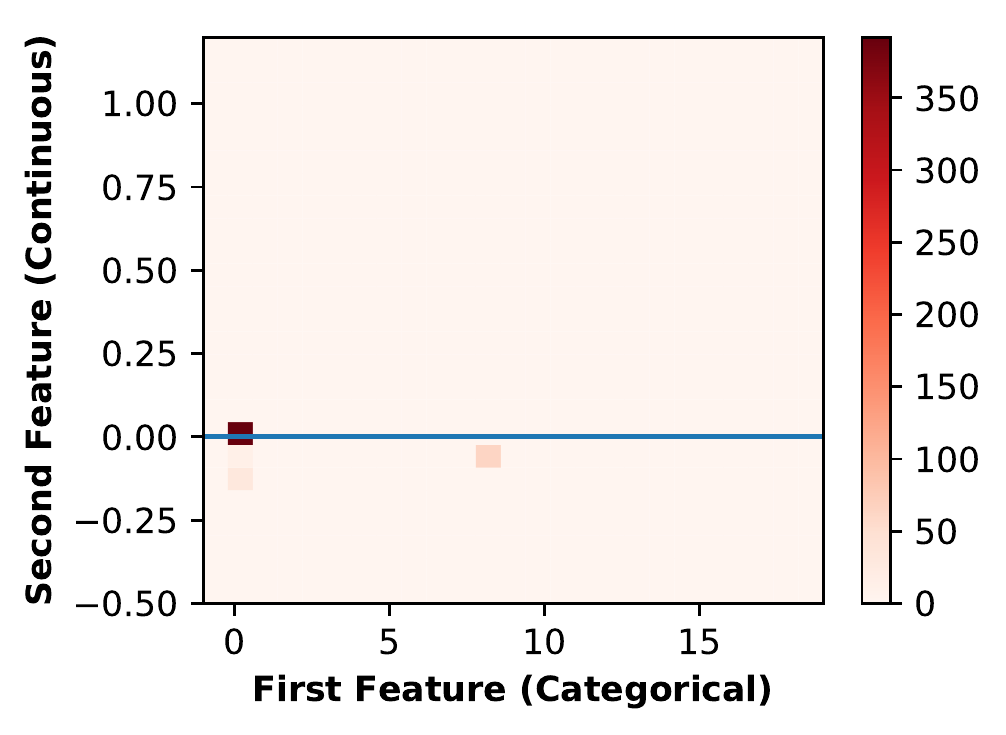}} 
\hfill
\subfigure[$E(\bm{x})$ by gs. \label{fig:discrete_laugel}]{\includegraphics[width=0.30\columnwidth]{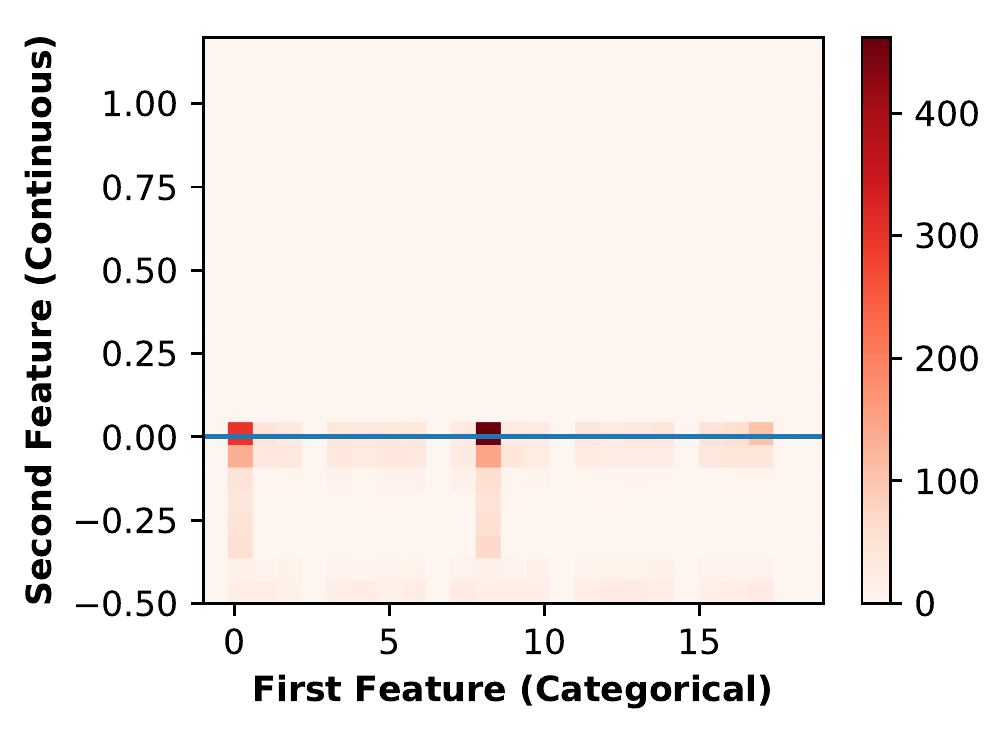}}
\hfill 
\subfigure[$E(\bm{x})$ by cchvae. \label{fig:discrete_cchvae}]{\includegraphics[width=0.30\columnwidth]{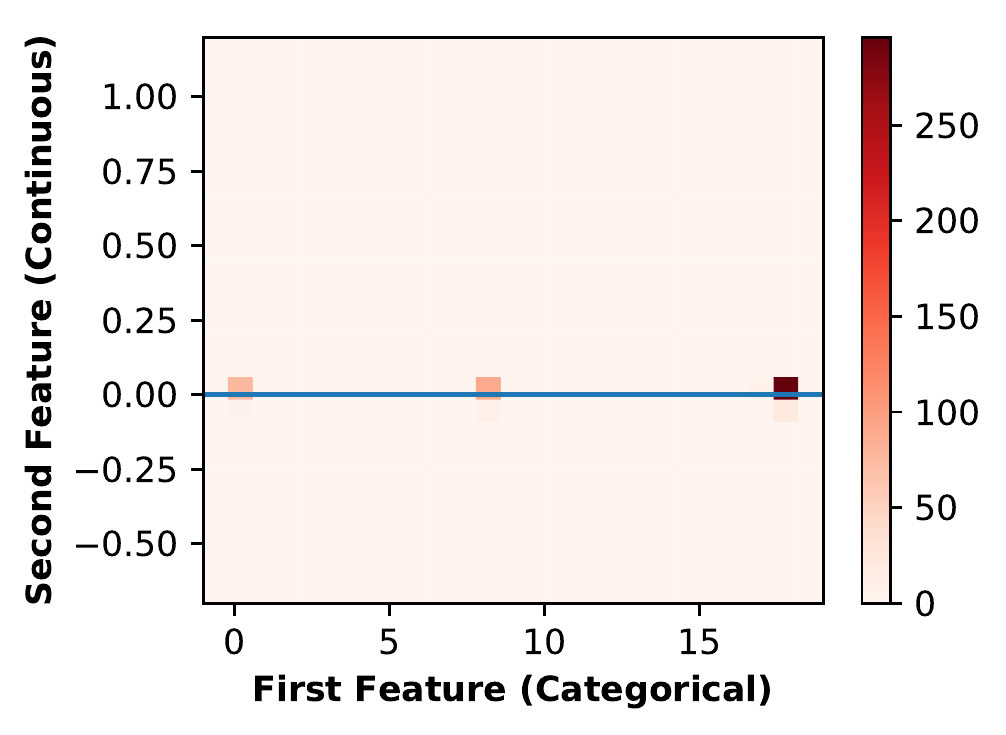}}%

\subfigure[Data generating process. \label{fig:dgp_discrete}]{\includegraphics[width=0.40\columnwidth]{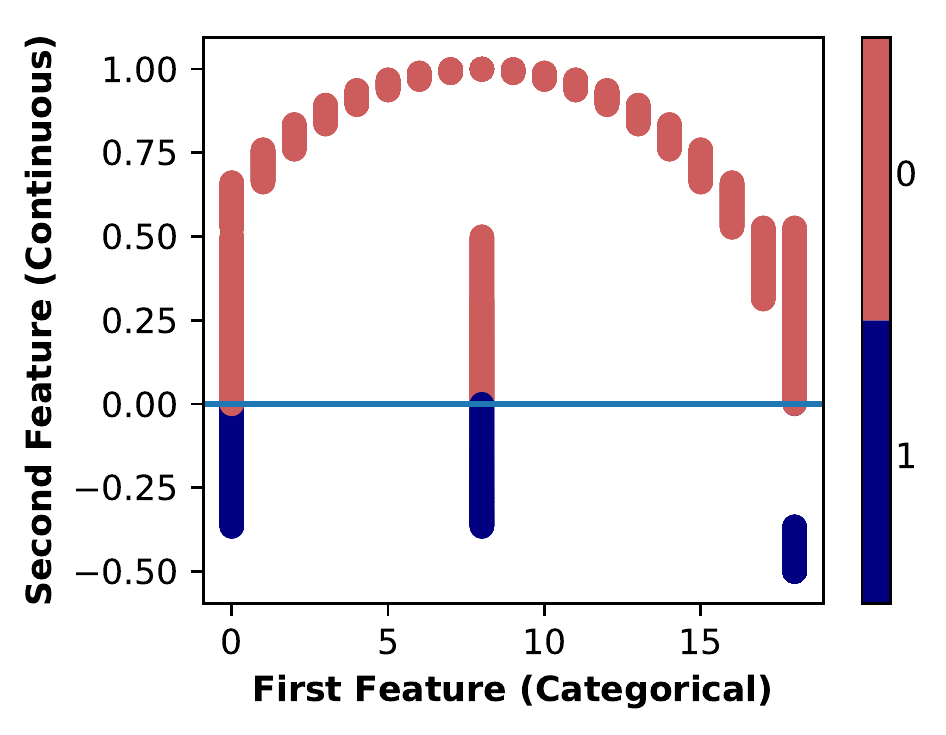}}%
\hfill
\subfigure[Test data cond. on $\Delta z$. \label{fig:discrete_delta}]{\includegraphics[width=0.40\columnwidth]{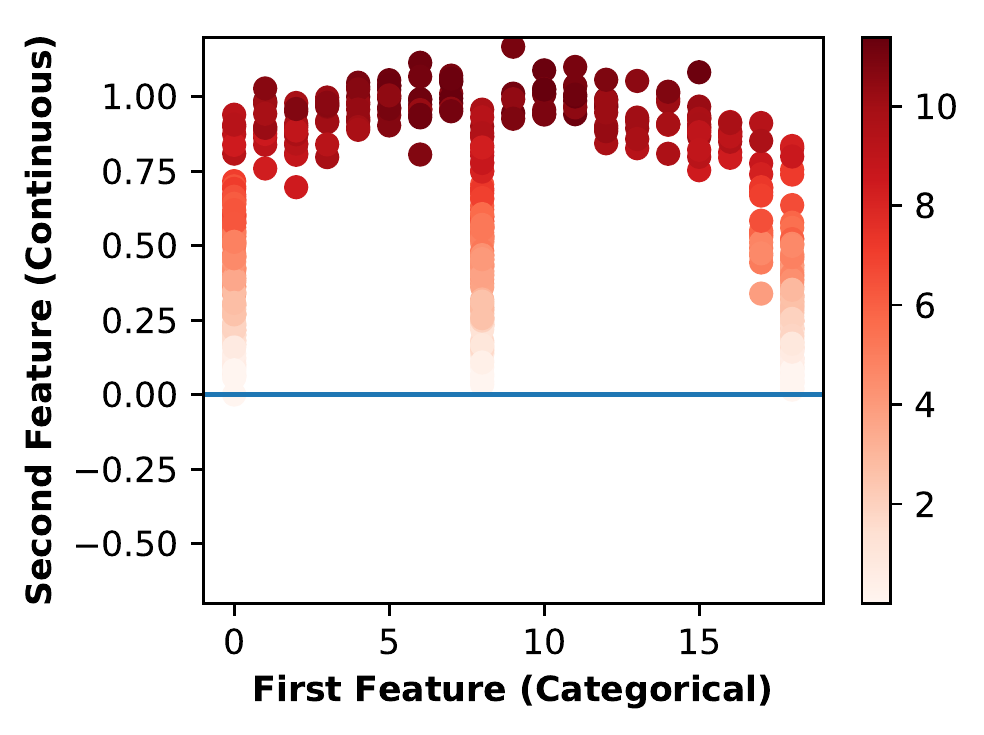}}

\caption{Example \ref{ex:make_moons}. Heterogeneous features. AR generates counterfactuals $E(\bm{x})$ from 1st and 9th category. GS generates counterfactuals from all categories. CCHVAE generates counterfactuals from 1st, 9th and 19th category. Figure \ref{fig:discrete_delta} shows test data conditional on $\Delta z = \tilde{z}^*-\hat{z}$, which indicates how much $\hat{z}$ needs to change to alter the prediction.}
\label{fig:Discrete}
\end{figure}

\paragraph{\textbf{Heterogeneous features}.}
Figures \ref{fig:dgp_discrete}, \ref{fig:dgp_2dhist_discrete} and \ref{fig:dgp_2dhist_test} depict the true data generating process with corresponding distribution of labels, corresponding 2d-histogram and test observations from the 0-class. Figure \ref{fig:discrete_recon} depicts the 2d-histogram from the reconstructed test data. Despite the simplistic class assignment, finding attainable counterfactuals might not be trivial in this case since the data density is very fragmented.


Next, figures \ref{fig:discrete_ar}-\ref{fig:discrete_cchvae} depict 25 counterfactuals generated using AR, HCLS and our C-CHVAE, respectively. For AR, counterfactuals appear in the 1st and the 9th category. For GS, counterfactuals appear in all categories. For our C-CHVAE, counterfactuals appear in the 1st, 9th and 19th category. The model correctly produces counterfactuals for all categories.


\section{Data}\label{sec:appendix_data}
\subsection{Real world example: ``\texttt{Give Me Some Credit}''}
In the following, we list the specified pretrained classification models as well as the parameter specification used for the experiments. We use 80 percent of the data as our training set and the remaining part is used as the holdout test set. Additionally, we allow $f$ access to all features, i.e. $f(\bm{x}^f, \bm{x}^p)$. The state of features can be found in table \ref{tab:state_givme}.

\paragraph{\textbf{AR} \citep{spangher2018actionable}.}
The AR algorithm requires to choose both an action set and free and immutable features. The implementation can be found here: \url{https://github.com/ustunb/actionable-recourse}. We specify that the \emph{DebtRatio} feature can only move downward \citep{spangher2018actionable}. The AR implementation has a default decision boundary at $0$ and therefore one needs to shift the boundary. We choose $p_{AR} = 0.50$, adjusting the boundary appropriately. Finally, we set the linear programming optimizer to \emph{cbc}, which is based on an open-access \texttt{python} implementation. As $f$, we choose the $l_2$-regularized logistic regression model.

\paragraph{\textbf{GS} \citep{laugel2017inverse}.}
GS is based on a version of the YPHL algorithm described above. As such we have to choose appropriate step sizes in our implementation to generate new observations from the sphere around $\bm{x}$. We choose a step size of 0.1. As $f$, we choose the $l_2$-regularized logistic regression model.

\paragraph{\textbf{HCLS} \citep{lash2017generalized}.} In our experiment we used their baseline \texttt{MATLAB} implementation, which can be found here: \url{github.com/michael-lash/BCIC}. HCLS requires us to choose a budget, which we set to 10. It also requires to choose a cost associated with changing each feature. We set it equal to $1$ for all features. As $f$, we choose SVM with the Gaussian kernel, which delivered good results in reasonable time. Initially, we tried to choose the linear kernel, but after training for several hours with no convergence, we decided against it. We also experimented with different standardization forms (minmax standardization, z-score standardization), which did not help. For the evaluation metric, we choose accuracy and we used a \emph{balance} option that weighs each individual sample inversely proportional to class frequencies in the training data. We had to specify an indirectly changeable feature, which we set to \emph{NumberOfTimes90daysLate}. Finally, we had to choose the direction (Dir.(HCLS) in table \ref{tab:state_givme}) in which \emph{every} free feature is allowed to move.

\paragraph{\textbf{C-CHVAE} (ours).}
For our algorithm we made the following choices. We set the latent space dimension of both $\bm{s}$ and $\bm{z}$ to 5 and 6, respectively. For training, we used 50 epochs. Table \ref{tab:state_givme} gives details about the chosen likelihood model for each feature. For count features, we use the Poisson likelihood model, while for features with a support on the positive part of the real line we choose log normal distributions. As $f$, we choose the $l_2$-regularized logistic regression model.


\begin{table}
    \centering
     \begin{adjustbox}{max width=\columnwidth}
    \begin{tabular}{lccc}
    \toprule
         Feature & Free & Model & Dir. (HCLS) \\
         \cmidrule(lr){1-4}
         \emph{Revolving Utilization Of Unsecured Lines} & Y & log Normal & $\downarrow$ \\ 
         \emph{Age} & N & Poisson & \\
         \emph{Number Of Times 30-59 Days Past Due Not Worse} & Y & Poisson & $\downarrow$  \\
         \emph{Debt Ratio} & Y & log Normal & $\downarrow$  \\
         \emph{Monthly Income} & Y & log Normal & $\uparrow$ \\
         \emph{Number Open Credit Lines And Loans} & Y & Poisson & $\downarrow$ \\
         \emph{Number Of Times 90 days Late} & Y & Poisson & indirect  \\
         \emph{Number Real Estate Loans Or Lines} & Y & Poisson & $\downarrow$ \\
         \emph{Number Of Times 60-89 Days Past Due Not Worse} & Y & Poisson & $\downarrow$  \\
         \emph{Number Of Dependents} & N & Poisson & \\
         \bottomrule
    \end{tabular}
    \end{adjustbox}
    \caption{``\texttt{Give Me Some Credit}'': State of features and likelihood models.}
    \label{tab:state_givme}
\end{table}

\begin{table}
    \centering
     \begin{adjustbox}{max width=\columnwidth}
    \begin{tabular}{lccc}
    \toprule
         Feature & Free & Model & Dir. (HCLS) \\
         \cmidrule(lr){1-4}
         \emph{MSinceOldestTradeOpen} & N & Poisson &  \\
         \emph{AverageMInFile} & N & log Normal & \\ 
         \emph{NumSatisfactoryTrades} & Y & Poissonl & $\uparrow$  \\
         \emph{NumTrades60Ever/DerogPubRec} & Y & log Normal & $\downarrow$  \\
         \emph{NumTrades90Ever/DerogPubRec} & Y & log Normal & indirect \\
         \emph{NumTotalTrades} & Y & Poisson & $\downarrow$ \\
         \emph{PercentInstallTrades} & Y & log Normal & $\uparrow$ \\
         \emph{MSinceMostRecentInqexcl7days} & Y & Poisson & $\downarrow$ \\
         \emph{NumInqLast6M} & Y & Poisson & $\downarrow$ \\
         \emph{NetFractionRevolvingBurden} & Y & log Normal & $\downarrow$ \\
         \emph{NumRevolvingTradesWBalance} & Y & Poisson & $\uparrow$ \\
         \emph{NumBank/NatlTradesWHighUtilization} & Y & log Normal & $\uparrow$ \\
         \emph{ExternalRiskEstimate} & N & log Normal &   \\
          \emph{MPercentTradesNeverDelq} & Y & log Normal & $\downarrow$ \\
          \emph{MaxDelq2PublicRecLast12M} & Y & Poisson & $\downarrow$ \\
          \emph{MaxDelqEver} & Y & Poisson & $\downarrow$ \\
          \emph{NumTradesOpeninLast12M} & Y & Poisson & $\downarrow$ \\
          \emph{NumInqLast6Mexcl7days} & Y & Poisson & $\downarrow$  \\
          \emph{NetFractionRevolvingBurden} & Y & Poisson & $\downarrow$ \\
          \emph{NumInstallTradesWBalance} & Y & Poisson & $\uparrow$ \\
          \emph{NumBank2NatlTradesWHighUtilization} & Y & Poisson & $\downarrow$ \\
          \emph{PercentTradesWBalance} & Y & log Normal &  $\uparrow$ \\
         \bottomrule
    \end{tabular}
    \end{adjustbox}
    \caption{\texttt{HELOC}: State of features and likelihood models.}
    \label{tab:state_default}
\end{table}

\subsection{Real world example: \texttt{HELOC}}


The \emph{Home Equity Line of Credit (HELOC)} data set consists of credit applications made by homeowners in the US, which can be obtained from the FICO community.\footnote{\url{https://community.fico.com/s/explainable-machine-learning-challenge?tabset-3158a=2}.} The task is to use the applicant's information within the credit report to predict whether they will repay the HELOC account within 2 years. Table \ref{tab:state_default} gives an overview of the available features and the corresponding assumed likelihood models.

\paragraph{\textbf{AR and GS}} As before. Additionally, we do not specify how features have to move.
\paragraph{\textbf{HCLS}} As $f$, we choose SVM with the linear kernel. We specified \emph{NumTrades90Ever/DerogPubRec} as the indirect feature. Again, we had to specfiy which directions features move, which we indicated in the 'Direction' column of table \ref{tab:state_default}.
\paragraph{\textbf{C-CHAVE} (ours)}
For our algorithm we made the following choices. We set the latent space dimension of both $\bm{s}$ and $\bm{z}$ to 1 and 10, respectively. For training, we used 60 epochs. Table \ref{tab:state_default} gives details about the chosen likelihood model for each feature. The rest remains as before.

\end{document}